# Behavior of Hyper-Parameters for Selected Machine Learning Algorithms: An Empirical Investigation

Anwesha Bhattacharyya, Joel Vaughan, and Vijayan N. Nair
Corporate Model Risk
Wells Fargo Bank N.A.

Hyper-parameters (HPs) are an important part of machine learning (ML) model development and can greatly influence performance. This paper studies their behavior for three algorithms: Extreme Gradient Boosting (XGB), Random Forest (RF), and Feedforward Neural Network (FFNN) with structured data. Our empirical investigation examines the qualitative behavior of model performance as the HPs vary, quantifies the importance of each HP for different ML algorithms, and stability of the performance near the optimal region. Based on the findings, we propose a set of guidelines for efficient HP tuning by reducing the search space.

## 1. Introduction and Overview

Most machine learning (ML) algorithms have hyper-parameters (HPs) that have to be tuned (optimized) as part of the model training process. HP Optimization (HPO) can be a time-consuming process, and this is especially so for algorithms with a large number of HPs. This is often a limiting factor in the use of ML algorithms, especially when the number of HPs is large and the dataset is big. The goal of this paper is to study the behavior of HPs and use the findings to develop guidelines that can reduce the complexity of the HPO process for the algorithms in our study.

We focus on three supervised ML (SML) algorithms: **R**andom **F**orest (RF), E**X**treme **G**radient **B**oosting (XGB), and **F**eed **F**orward **N**eural **N**etwork (FFNN). These are the three most commonly used algorithms with structured data and are known to have good predictive performance. See (Liu, Hu, Chen, & Nair, 2022) and references therein.

1) RF, proposed by (Breiman, 2001), is an example of ensemble learning that combines information from multiple trees (base learners) to improve performance. Given a dataset, it creates multiple subsets by bootstrapping the observations (rows), grows trees on each subset, fits piecewise constant models on each tree, and averages them (bootstrap aggregation or bagging). In addition, to de-correlate the base model predictions and achieve a larger amount of variance reduction, each tree split is based on a random subset of the features (columns). There are several HPs associated with RF, such as number of trees, depth of the trees, and so on. See Section 2 for additional HPs.
2) Gradient Boosting Machines (Friedman., 2001) is another ensemble algorithm. It is a sequential procedure that starts with a weak learner (typically piecewise constant trees) and repeatedly applies boosting to reduce the errors in model fitting. In this paper, we consider XGB (Guestrin, 2016), an efficient implementation of gradient boosting, which is distributed and regularized. It also relies on HPs such as depth and number of trees. See Section 2 for additional HPs.

3) Neural networks have a longer history and were originally developed to mimic biological networks (see, for example, (Goodfellow, Bengio, & Courville, 2016)for the history and extensive discussion). FFNNs consist of an input layer, and output layer, and several hidden layers with multiple neurons. They are fully connected (each neuron in one hidden layer is connected to every other neuron in the next hidden layer and the connections are only in the forward direction. In our study, we restricted attention to FFNNs with one and two hidden layers, ReLU activation functions, and utilized Keras models built on Tensor Flow for model fitting. FFNNs have more HPs than the tree ensemble algorithms, and a list is given in Section 2.

The predictive performances achieved by these ML algorithms depend on the settings of the HPs, so appropriately tuning the parameters is key to achieving good results [see (Weerts, Mueller, & Vanschoren, 2020) and (Zhang, et al., 2021)]. A recent survey of HPO techniques can be found in Hu et al. (2021) covering both batch and sequential methods. Grid search is known to be an inefficient technique in this problem. Random search over the HP grid is the preferred solution when the number of HPs is not large. However, random search considers only a subset of the parameter configurations and may miss the best value or even region. Sequential algorithms such as Bayesian Optimization with Gaussian Processes and Tree Structured Parzen estimators or mixed ones such as Hyperband are typically used with FFNNs that have many layers ore other deep NNs. All of these search algorithms are computationally resource-intensive when the dataset is very large or the number of HPs is large (as is the case with most DNNs).

HPO is often treated as a stand-alone exercise and a necessary evil. There is limited effort on developing insights into the following key questions:

- How does the performance vary as a function of the HPs?
- Are there some parameters that are always more important than others?
- Are there multiple high performing regions?
- Is performance stable around the optimal parameter settings? and
- How would the optimal settings change if there are perturbations in the data?

The goal of this paper is to address some of these questions through an empirical investigation. As we will see, these insights provide guidelines on how to conduct HPO more efficiently and develop a multi-stage search strategy.

There is some previous work along these lines. A functional ANOVA technique was exploited in (Hutter, Hoos, & Leyton-Brown, 2014) in assessing the individual importance of HPs and their interactions for RF based on results obtained from Bayesian optimization. (Rijn & Hutter, 2018) used the same technique to evaluate the performance of selected HPs for RF, AdaBoost and SVM. (Jin, 2022) also looked at importance of HP from a risk reduction perspective using a subsampling-based approach and evaluated selected HPs of XGBoost in their implementation.

This empirical investigation is based on three real datasets and one SIM dataset, covering binary and continuous responses. For each, we considered a large parameter space for the relevant parameters and did a grid search over all possible combinations of HPs. The number of combinations varied with datasets and algorithms (see

Table 5), but they ranged from around 1,500 to about 7,000. Specifically, for each dataset, we trained the three algorithms at each possible HP combination and evaluated the predictive performance on a hold-out dataset, hereafter referred to as Validation (Valid) data. The performance measures were analyzed to answer the questions above and related ones.

The rest of the paper is organized as follows. Section 2 provides a description of the datasets used, the HPs optimized, and the total configurations evaluated for each model. Section 3 provides a high-level summary of the findings from the study as well as a description of two-stage search strategies to reduce the computational burden. These strategies are informed by the empirical findings and lead to reduced computational burden. Sections 4, 5 and 6 describe the analysis of the performance metrics (AUC/Logloss/MSE) across the HP configurations for each dataset of the RF, XGB and FFNN models respectively. For each model, we carried out a global analysis using ANOVA techniques to systematically evaluate the importance of the HPs and their interactions. There are HPs that have global impact and must be tuned to achieve good model performances. We also conducted an analysis on the local behavior of the HPs by looking at the high performing models. This has illuminated several weak interactions between HPs. As a result, there can be multiple models with equally competitive performance. We also explored issues like overfitting. Section 7 describes the consistency between the performance metrics for binary response datasets. We end our report with concluding remarks in Section 8.

## 2. Description of Datasets and Hyper-parameters

### 2.1. Datasets

Our analysis is based on three real datasets: Home Lending Data, Personal Lines and Loans Data, and the Bike Share Data. In addition, to further study the FFNN algorithm, we used a SIM dataset with continuous responses.

Table 1: Description of datasets

| Name | Abbreviation | Type of Response | Number of variables | Dataset Size – Training: Validation |
|---|---|---|---|---|
| Home Lending | HL | Binary | 62 | 333,431: 333,706 |
| Personal Line of Loans | PLL | Binary | 143 | 35,583: 26,687 |
| Bike Share | BS | Continuous | 11 | 9,384: 3,997 |
| Simulation (see model form below) | SIM | Continuous | 14 | 30,000: 10,000 |

#### a) Home Lending Data

The Home Lending dataset deals with residential mortgages, and the response is an indicator (= 1) for “troubled loans”. The original dataset was huge, and we used a randomly selected subset of one million for our analysis. There were more than 90 variables out of which we selected 62 predictors, omitting some transformed covariates based on subject matter expertise.

#### b) Personal Line of Loans Data

This dataset deals with default on unsecured personal line of credit. The original dataset had 90K observations with 500 variables. The response in a binary variable indicating default (= 1) or not. We did some initial screening and selected 143 variables.

**c) Bike Share Data**

This is a publicly available data obtained from the UCI repository (UCI - Bike Share). It has hourly bike rentals for the years 2011 and 2012 from Capital Bike Share system. There are a total of 17,379 records with hourly bike rentals along with relevant weather information. We used the log of the count as response.

**d) Simulated Data**

The simulated dataset has continuous responses and was based on the following model:

$$f(\boldsymbol{x}) = |x_0| + 0.5x_1 + -2\log(|x_2|+1) + \exp(0.5x_3) + (0.5|x_4|+1)^{-1} + 1.5x_0x_1 + |x_0\,x_1x_2| + 1.5\log(|x_2+x_3+x_4|+1) + 0.75\max(x_3,x_4) + \exp[0.35(x_4-x_2)] + 0.7[|x_5| + 0.5x_6 - 2\log(|x_7|+1) + \exp(0.5x_8) + (0.5|x_9|+1)^{-1} + 1.5x_5x_6 + |x_5x_6x_7| + 1.5\log(|x_7+x_8+x_9|+1) + 0.75\max(x_8,x_9) + \exp[0.35(x_9-x_7)]\,] + 0.4\left[|x_{10}| + 0.5x_{11} + -2\log(|x_{12}|+1) + \exp(0.5x_{13}) + (0.5|x_{14}|+1)^{-1} + 1.5x_{10}x_{11} + |x_{10}x_{11}x_{12}| + 1.5\log(|x_{12}+x_{13}+x_{14}|+1) + 0.75\max(x_{13},x_{14}) + \exp[0.35(x_{14}-x_{12})]\right].$$

The response was $y = f(\boldsymbol{x}) + \epsilon$, where $\epsilon \sim N(0,1)$. We used a block correlation structure for the predictors in groups of two and three consecutively with correlation.

All the algorithms were developed on training subsets, and model performances were evaluated on the validation subsets. For performance metrics, we used mean square error (MSE) for continuous responses and both Logloss and AUC for binary responses.

### 2.2. Hyper-parameters and Configurations

The HPs studied for each of these algorithms are listed in Table 2-Table 4. Other HPs involved in model building were held fixed at their default values as specified by the respective API.

Table 2: HPs for RF

| HP | Description and Background |
|---|---|
| Depth | Depth or number of layers of each tree. This parameter controls the complexity of the individual trees. |
| Trees | Number of trees in the forest: control robustness of predictions. |
| MSL: Minimum samples per leaf | Minimum number of samples (observations) at each leaf node. This parameter helps to control over-fitting. |
| Max_p: (column sample) | Maximum number of features (variables) used to build a tree. If $p$ denotes the total number of features in the dataset, (Hastie, Tibshirani, & Friedman, 2009) suggest that Max_p should be about the order of $\sqrt{p}$ for binary responses and $p/3$ for continuous responses. |

Row sampling (proportion of rows) is typically varied with RF, but we did not study it here as an HP. Sklearn implementation sets the default value of row sampling ratio at 1.

Table 3: HPs for XGB

| HP | Description |
|---|---|
| Lr_rate: Learning rate<br>Range: [0,1] | Step size shrinkage used in update to prevent overfitting.<br>After each boosting step it shrinks the feature weights to make the boosting process more conservative. |
| Trees | Number of estimators (or boosting iterations) |
| Depth | Maximum allowed depth of each tree |
| Alpha - L1 | L1 regularization on weights. |
| Lambda - L2 | L2 regularization on weights. |

Row and column sample ratios are usually not tuned when training XGB model, so we have not included them as HPs in our extensive study. However, we carried out a side study to evaluate the effect of column sampling and minimum child weight. The results reported in the Appendix.

Table 4: HPs for FFNN

| HP | Description |
|---|---|
| Lr_rate: Learning rate | Learning rate in the Keras ADAM optimizer |
| Batch_size | The number of observations in each batch for computing gradients |
| Layer1 | Number of neurons in Layer 1 |
| Layer2 | Number of neurons in Layer 2 |
| L1 | L1 value to the L1_L2 penalty in Keras layers |
| L2 | L2 value to the L1_L2 penalty in Keras layers |
| Dropout | A form of regularization added to the hidden layers |

We restricted attention to ReLU activation functions for the internal (hidden) layers. For the final (output layer), we used the sigmoid function for binary regression and linear function for continuous regression. The discrete variables were one-hot encoded, and the input was min-max scaled. We used batch normalization for each layer and utilized mini-batch shuffling for training the model. Additionally, we trained the model using early stopping and a large number of potential epochs (1000).

Table 5: Number of models (HP configurations) evaluated

| Algorithm | HL | PLL | BS | SIM |
|---|---|---|---|---|
| XGB | 6,250 | 6,250 | 6,250 | NA |
| RF | 3,600 | 1,960 | 1,512 | NA |
| FFNN | 5,184 | 6,912 | 6,912 | 6,912 |

For each dataset, we built the models on an extensive grid of HPs in order to understand their impact. The grid points differed slightly across datasets and were selected after some initial experiments on each dataset. In particular, choices of HPs for minimum samples per leaf, maximum features, or batch size varied with size of the datasets. The total number of models built in each case is given in Table 5.

## 3. Summary of Findings and Reduced Search Strategy for Tuning HPs

### 3.1. Background on Analyses

Our findings in the next three sub-sections are based on the following types of analyses:

a) We conducted a global analysis across the large numbers of HP configurations in Table 5 to identify the important HPs for each algorithm in terms of main effects and any interactions. This was done using an analysis of variance (ANOVA) with the performance metrics as responses and the HPs as factors. Since the underlying assumptions needed for ANOVA and F- tests were unlikely to hold, we used the results informally and just ranked the effect sizes based on the percentage contribution of their F-statistic ($\frac{F_{effect}}{\sum_{effect} F_{effect}} \times 100$). We also visualized the pairwise interactions through contour plots. We will see that these analyses provided useful insights into the relative importance of the different HPs and their interactions.

b) Next, we restricted attention to high-performing models and studied the behavior of the HPs within these models. These helped to provide additional information on local behaviors, especially local interactions, which were not apparent in the global analyses.

c) Finally, we studied the robustness of the best fitting model by examining the gap between training and validation subset defined as

$$metric_{gap} = |metric_{train} - metric_{valid}| .$$

A large gap indicates that the model was possibly over-fitting the training subset. We compared the gap of the best-fitting model with that of an alternative high-performing model. In addition, we also conducted a limited robustness study by comparing the performances of both models under small levels of perturbations of the predictors.

In addition, we propose two-stage strategies for HP tuning for each algorithm. This is based on our findings and ensures that practitioners do not have to start from ground zero each time they have to do HPO for these algorithms. We describe results of the proposed strategies for different datasets and show that the optimal model recovered from the reduced search strategies are consistently in the upper 1% of the total set of models originally evaluated.

### 3.2. Random Forest

#### 3.2.1 Findings

**i) Global Behavior:**

- Of the five HPs, Depth was the most important, and optimal models preferred deep trees (Depth > 11). However, higher values of Depth without any form of regularization lead to over-fitting.
- MSL was the second most important and helped to control overfitting effectively.
- As long as Max_p was larger than $\sqrt{p}$, its impact was not strong. However, if number of features in a dataset was low to begin with, restricting Max_p to be around $\sqrt{p}$ or lower can result in under-fitting.
- As long as the value for Tree is moderately large($> 100$), average performance did not vary with Trees.

ii) **Local Behavior:** This analysis is based on HP behavior when we focused on the top 50 models.

- There were some interactions between Max_p and both Depth and MSL. However, they were weak. Nonetheless, such interactions may lead to multiple models with similar performance.
- Due to the presence of such weak interactions, it is possible to gain some more accuracy by tuning Max_p in addition to tuning Depth and MSL.

iii) **Robustness:** These conclusions are based on examining the gap between performances with training and validation subset as well as a robustness analysis based on local perturbations of the predictors.

- For the BS data, the best performing model had a relatively large gap between performances in validation and training subset compared to an alternative good-performing model. This can be caused by overfitting the training subset. However, the robustness analysis based on perturbations of the predictors did not identify any lack of robustness of the best performing model. Of course, this conclusion is based on just one dataset. Nevertheless, it is a cautionary note on relying just on the gap statistic to judge robustness.

### 3.2.2. Two-Stage HP Tuning Strategy

Stage 1:

a) Recall that the effect of Max_p is not strong as long as it is at or above $\sqrt{p}$. So we fix it at around $\sqrt{p}$.
b) Further, Trees did not have a major effect as long as it was between 300 and 500 suffices. So we fix it to be between these values.
c) Depth was the most important HP, and top-performing models had Depth$> 10$. So, for this HP, we take the grid to be between 10 and 20.
d) MSL was also important but good models usually had MSL below $\sqrt{n}$ where $n$ is the training subset size. So we take the grid for MLS to be geometrically spaced values from 1 to$\sqrt{n}$ .
e) Conduct a two-dimensional search over Depth and MSL, with the others set at the prescribed values, to identify $Depth_{r-opt}$ $and$ $MSL_{r-opt}$, where `r-opt' stands for best in the reduced search space. This is just a two-dimensional tuning problem and can be conducted using random search with a fine grid, thus taking into the weak interaction between MSL and Depth.

Stage 2: Given$Depth_{r-opt}$ and $MSL_{r-opt}$,

a) Tune Max_p. Due to its mild interaction with Depth and MSL, Max_p will automatically pick the best value suited for the chosen values of Depth and MSL when tuned. We suggest taking values of Max_p as integer multiples of$\sqrt{p}$. Since this is just a one-dimensional problem, grid-search will be the simplest. Note that due to the interactions between Max_p with Depth and MSL, it is possible to obtain suboptimal performance by tuning Max_p separately. However, our evaluation of the strategy on the three datasets showed that this relative decrease in performance was small (less than 0.1%).

b) Tuning the number of Trees along with Max_p is recommended as optional, and we did not do it in our investigation reported below.

We studied the performance of the two-stage approach with Trees = 300 and Max_p = $\sqrt{p}$ set at Stage 1 and Trees = 300 in Stage 2. Table 6 reports performance results comparing the best model from the reduced search with those of all models from the extensive grid search. The relative decrease in performance is defined as $\frac{|metric_{g-opt} - metric_{r-opt}|}{metric_{g-opt}} \times 100$, where *g-opt* is the global best model from the extensive grid search and *r-opt* is the best one from the reduced search. We see from Table 6 that *r-opt* was consistently in the top 1% of all models in the global search. The boxplots in Appendix A.a provide a more detailed comparison. In summary, the two-stage strategy based on the findings are useful in reducing the computational burden with very little loss in performance.

Table 6: Results of Proposed Strategy for RF models

| Datasets | Metric | Total number of models in original grid search | Percentage of models having higher performance | Relative decrease in performance |
|---|---|---|---|---|
| HL | AUC | 3,600 | 0.17% | 0.05% |
| | Logloss | 3,600 | 0.03% | 0.06% |
| PLL | AUC | 1,952 | 0.2% | 0.07% |
| | Logloss | 1,952 | 0.92% | 0.1% |
| BS | MSE | 1,512 | best | 0% |

### 3.3. XGBoost

#### 3.3.1 Findings

- **Global Behavior:** Depth, Trees, and Lr_rate were the most important HPs.
    - XGB models preferred shallower trees (Depth $\leq$ 7) compared to RF models.
    - Extremely low Lr_rate (<=0.01) can lead to poor performance unless balanced by a very large number of trees (> 500). Obviously, this would increase model training time.
- Depth, Trees, and Lr_rate interacted with each other, and the interaction between Lr_rate and Trees was especially strong. It appears that the best way to tune the value for Trees is to use early stopping.

- Top models often had the highest possible value for regularization HPs, alpha (L1) and lambda (L2), indicating that higher penalty values might be required for good performance.
- L1 had higher impact than L2 except for the Bike Share dataset with continuous response, where the top models had low L1 penalty and high L2 values.
- Usually, models were regularized adequately through L1 and L2 penalty when the grid allowed for sufficient exploration. Additional regularization was provided by HPs like minimum child weight and column sampling, as seen in the extension study (see Appendix). However, they were impactful only when a model had not been regularized through L1 and L2 penalty sufficiently.

i) **Local Behavior:** As with RF case, we conducted an analysis of local HP behavior by restricting attention to the top 50 models.

- We observed similar pairwise interactions among Lr_rate, Trees, and Depth.
- In addition, there was significant negative rank correlation between the L1 and L2 penalties for HL and BS datasets. It appears that the top models preferred one or the other form of regularization but not both.
- There was positive correlation between Lr_rate and L1 and L2 penalties. High Lr_rate causes over-fitting and requires more regularization.
- Overall, the minor interactions which were not strong in the global analysis appear to be important in local behavior and explain the existence of multiple competing models.

iv) **Robustness:** As with the RF case, we examined the robustness of the best fitting model based on the gap between performances with training and validation subset and local perturbations of the predictors.

- The models that over-fitted usually had high values for one or more of these HPs: Depth, Lr_rate, and Trees. Either they had too low regularization penalties or required more regularization than currently allowed in our grid.
- Once again, the best fitting model for BS dataset had a relatively large gap value compared to an alternative well-performing model. We did not observe this behavior for the HL and PLL datasets with binary responses. A local perturbation analysis of the predictors confirmed that the best-fitting model exhibited some level of non-robustness. This suggests that we should be cautious in picking the best-fitting model without a robustness analysis for possible over-fitting.

### 3.3.2. Two-Stage HP Tuning Strategy

<u>Stage 1:</u>

a) Tune Depth and Lr_rate, two of the three important HPs, in Stage 1 and obtain the corresponding best setting of Trees through early stopping.
b) (Caruana & Niculescu-Mizil, 2005) showed that models with lower Lr_rate perform better, but extremely low Lr_rate can also lead to poor performance. Hence, limit Lr_rate to be between 0.02 and 0.1 and use a grid in log-scale so that we can search over more values in the lower range.
c) XGB models prefer low Depth($\leq$ 7), so set the range of Depth to be between 3 and 7. This range allows us to capture interactions with the other important HPs.

d) The interaction between Lr_rate and Trees is strong, so use early stopping to control it. (We used early stopping rounds = 20 in our investigation.) The maximum number of trees can be fixed to a large value like 500 or 1000. (We chose 500 in our investigation reported below.)
e) Keep L1 and L2 penalties at default values of 0 and 0.1.
f) Tune Depth and Lr_rate in the ranges described in items b) and c) to get the best configurations. This is a two-dimensional optimization problem, and one can use any of the search methods to get the best values of $Depth_{r-opt}$ $and$ $Lr_rate_{r-opt}$

Stage 2: Given $Depth_{r-opt}$ $and$ $Lr_rate_{r-opt}$:
a) Tune the penalty parameters for L1 and L2 from 0.001 to 15. Recall that best-fitting models often chose high values for these regularization HPs. We propose using a grid in log-scale allowing us to sample low values more frequently but also allow for the scope of higher regularization if required.
b) There are mild interactions between the regularization HPs with others. The tuning in the second stage will automatically choose best values for the chosen values of Depth and Lr_rate. However, due to the weak interaction of the regularization HP with Lr_rate and Depth, it is possible to obtain suboptimal performance in some cases. But our evaluation of the strategy showed that in case of the chosen datasets and existent full grid the decrease in relative performance was less than 2%.

The results in Table 7 show that the reduced search strategy often led to models that were the best in the global grid search and even in the worst cases they were in the top 1.5%. Thus, adopting the proposed strategy increases computational efficiency at very small loss in performance.

Table 7: Results of Proposed Strategy for XGB models

| Datasets | Metric | Total number of models in original grid search | Percentage of models having higher performance | Relative decrease in performance |
|---|---|---|---|---|
| HL | AUC | 6,250 | best | 0% |
| | Logloss | 6,250 | best | 0% |
| PLL | AUC | 6,250 | best | 0% |
| | Logloss | 6,250 | 0.016% | 0.03% |
| BS | MSE | 6,250 | 1.47% | 1.9% |

### 3.4. Feedforward Neural Network

Finally, we summarize the findings for FFNN. Note that the behaviors of the HPs were not always consistent across datasets, and in our summary, we make a note of findings that are general and findings that are specific to datasets.

**i) Global Behavior:**

- Dropout, L1 and size of Layer2 were usually the top parameters, with their relative importance varying with the datasets. Layer2 did not appear in the top 5 effects for PLL data and was included in the top 5 effects for the SIM data only through interaction with L1. L1 appeared in the top 5 effects in all datasets except for HL data.
- Higher dropout can lead led to diminished performance although this depended on interaction with Layer sizes and occurs when any one of the layers is relatively small.
- Models tended to over-fit when the size of either Layer1 or Layer2 was very large.
- Batch_size did not have a strong global effect on average performance as long as it is between 5% and 20% of the training subset.
- Lr_rate was more important with continuous responses (BS and SIM datasets) than with binary responses. However, in all cases, the total contributions of the F-statistics were less than 10%.
- Model performance can deteriorate as Lr_rate gets high. It seems advisable to select Lr_rate to be below 0.001 (the default Lr_rate in ADAM). In our experiments, we trained models with Lr_rate as low as 0.0005 and did not observe any negative impact as long as the models are allowed to train for sufficiently large number of epochs.
- Although L1 was an effective regularizer, it required careful tuning. When L1 values were on the higher side (> 0.001), any additional type of regularization can lead to under-fitting.
- There were some interactions between:
    - sizes of Layers and the regularization HPs – for example, Layer2 vs dropout in HL data and L1 vs Layer2 in SIM data,
    - within the regularization HPs – for example, L1 vs dropout and L1 vs L2 in PLL data.
    - However, these behaviors were not consistent across the datasets.

**ii) Local Behavior:** As with the RF and XGB cases, the findings below are based on the top 50 models. Overall, there were no consistent local patterns across the datasets for the HPs. Nevertheless, we note the following:

- The models seem to under-fit if all regularization HPs are high. Hence overfitting should ideally be controlled by tuning at most one or two regularization HPs.
- Sizes of Layer1 and Layer2 had positive interactions with at least one of the regularizing HP. For example, sizes of Layer1 and Layer2 have positive correlation with L2 in HL data, Layer1 size has positive correlation with dropout in PLL data, sizes of Layers 1 and 2 have positive correlation with L1 and L2 for BS and Sim data respectively.
- Lr_rate also had similar interactions. This suggests that when Lr_rate is high, higher regularization is needed to control over-fitting.
- With continuous response datasets, large batch size preferred L2 penalty.
- For the SIM dataset, all top 50 models had L1 penalty equal to 0 and were regularized through L2 and dropout. For this dataset, slight changes in L1 penalty when combined with any other forms of regularization lead to drastic change in performance.

iii) **Robustness:** Again, we examined the robustness of the best fitting model using the gap metrics and local perturbations of the predictors.

- The best performing model again had higher gaps for continuous datasets (BS and simulation) than binary cases, confirming the earlier findings with RF and XGB. However, unlike the earlier findings for local perturbation, the best fitting model was more robust than a competing alternative model. Thus, the robustness behavior seems to vary with datasets and algorithms. This reiterates the need for additional robustness analysis before deciding on the "best performing" HP configuration.

### 3.4.2. Two-Stage HP Tuning Strategy

Our general findings for the FFNN models are as follows

- There exists interaction between size of Layer1 and Layer2, interaction between the regularization HPs and interaction of Layer2 with the regularization. Hence there is no clear way of separating these HPs to formulate a two staged search strategy.
- We have seen Dropout having high impact as a regularizer across all datasets.
- Additionally, there is interaction between Lr_rate and regularizations HPs in top competing models.
- There is also strong interaction of batch size and regularization HPs in top models for continuous datasets.

Based on this information we propose the following two staged strategy for tuning FFNN models.

**Stage 1**

- Fix L1 and L2 at 0.
- Conduct search on the other 5 HPs simultaneously.
- Keep tuning range of Lr_rate <0.01, tune Batch_size between 5 to 20% of the data, use potential epoch of 1000 and train the model using early stopping.
- Obtain $Layer1_{r-opt}$, $Layer2_{r-opt}$, $Lr_rate_{r-opt}$ , $Batch_Size_{r-opt}$ in the reduced search space.

**Stage 2**

- In the second stage, tune L1 and L2 jointly given the optimum values of the other HPs in the reduced space and obtain $L1_{r-opt}, L2_{r-opt}$.

Note that due to high interaction between L1 and L2 and other HPs it is possible to obtain a suboptimal model. However, in most cases we observed that optimal models have low values of L1 and L2 penalty. Hence the decrease in performance by adopting the above two stage strategy should be small.

We have used this two-stage strategy on the extensive grid by emulating a random search on the already existing grid of models. The results are documented in Table 8. We observed that the resultant optimal model from this random search lies within top 1% models and the relative difference between the optimal model and the model from the alternate strategy is less than 1%. Hence using the alternate strategy usually results in a competing model. Joint tuning of all HP may result in an even better model.

Table 8: Results of Proposed Strategy for FFNN models

| Datasets | Metric | Total models in original grid search | Percentage of models having higher performance | Relative decrease in performance |
|---|---|---|---|---|
| HL | AUC | 5,184 | Best | 0% |
| | Logloss | 5,184 | Best | 0% |
| PLL | AUC | 6,912 | 0.42% | 0.66% |
| | Logloss | 6,912 | 0.95% | 0.44% |
| BS | MSE | 6,912 | Best | 0% |
| SIM | MSE | 6,912 | Best | 0% |

### 3.5. Other Findings

- For binary responses, we considered two metrics: AUC and Logloss. While there was general concordance among the two, the association was not exactly linear, and the two metrics can lead to different optimal models.
- For FFNN models, we observed that:
  - When a model is heavily regularized, the effects of the predictors can be diluted leaving us with essentially an intercept model.
  - For highly imbalanced data with many zeros, AUC is better able to discriminate among the poor performing models than Logloss.

## 4. Detailed Results in Support of the Findings for RF

We considered the following four HPs for RF implementation in Scikit-learn. In our grid, for MSL, we opted for geometrically spaced points starting from 1 to a multiple of $\sqrt{n}$ where $n$ is the number of observations in the training subset. Other HPs were set at default values. We trained the RF models at all possible configurations of the specified grid and recorded the values of the evaluation metric in both training and validation subsets for all the configurations.

Table 9: Random Forest HP grid specification

| HP | HL (n = 300K, p = 62 ) | PLL (n = 35K, p = 143 ) | BS (n = 9K , p = 11 ) |
|---|---|---|---|
| Depth | 3, 5, 7, … ,17 | 3, 5, 7, … ,17 | 3, 5, 7, … ,17 |
| Trees | 100,150,200, …, 500 | 100,200 ,…, 500 | 100,150,200, …, 500 |
| Max_p (multiples of $\sqrt{\boldsymbol{p}}$) | 7, 14,21,28,35 | 11, 22, 33, … , 77 | 3, 6, 9 |
| MSL | 1, 2, 5, 14, 34, 83, 202, 491, 1191, 2887 | 1, 2, 5, 13, 32, 78, 188 | 1, 2, 4, 9, 21, 45, 96 |

### 4.1. Global Analysis to Identify Important Hyper-Parameters

As noted in Section 3.1, we used ANOVA informally to do a global assessment of the HPs and their interactions by ranking the effect sizes. Table 10-Table 14 show the top five effects for each dataset based on Logloss and AUC metrics for binary response and MSE for continuous response. Here are our findings:

i) In most cases, Depth was the most important parameter. Trees with Depth >= 11 performed better.
ii) MSL was a close second. Good values for MSL were higher for binary response than for continuous responses (BS dataset).
iii) Max_p was important in some cases. The performance change was not as strong as with Depth and MSL.
iv) Trees (number of estimators) was not important, at least in our grid which had a minimum value of 100 Trees.
v) None of the interactions among the HPs was very important.

Table 10: ANOVA Table for HL Data
Top 5 effects by AUC

| HP | % F | %cum F |
|---|---|---|
| Depth | 92.59 | 92.59 |
| MSL | 5.78 | 98.37 |
| Max_p | 0.87 | 99.24 |
| Depth : MSL | 0.32 | 99.56 |
| Trees | 0.24 | 99.80 |

Table 11: ANOVA Table for HL Data
Top 5 effects by Logloss

| HP | % F | %cum F |
|---|---|---|
| Depth | 78.88 | 78.88 |
| MSL | 14.29 | 93.17 |
| Max_p | 5.89 | 99.06 |
| Depth: Max_p | 0.36 | 99.42 |
| Depth : MSL | 0.36 | 99.78 |

Table 12: ANOVA Table for PLL Data
Top 5 effects by AUC

| HP | % F | %cum F |
|---|---|---|
| MSL | 44.42 | 44.42 |
| Depth | 42.74 | 87.16 |
| Max_p | 8.2 | 95.36 |
| Trees | 3.12 | 98.48 |
| Depth : MSL | 1.11 | 99.59 |

Table 13: ANOVA Table for PLL Data
Top 5 effects by Logloss

| HP | % F | %cum F |
|---|---|---|
| Depth | 60.20 | 60.20 |
| MSL | 36.61 | 96.81 |
| Trees | 1.34 | 98.15 |
| Depth : MSL | 0.64 | 98.79 |
| Depth: Max_p | 0.39 | 99.18 |

Table 14: ANOVA Table for BS Data
Top 5 effects by MSE

| HP | % F | %cum F |
|---|---|---|
| Depth | 47.05 | 47.05 |
| Max_p | 42.11 | 89.16 |
| MSL | 9.88 | 99.04 |
| Depth : MSL | 0.42 | 99.46 |
| Depth : Max_p | 0.28 | 99.74 |

We next examine the effects and interactions visually. Specifically, we considered the three pairs Depth, MSL, and Max_p. For a fixed pair of these HPs, we computed the average performance of the models across the other HP configurations. The results are displayed as contour plots in Figure 1. To reiterate, the figure shows average performance patterns for a pair of HPs. For given dataset, the contour plots are mapped to the same color scale across different HP pairs for ease of comparison. In case of BS data, instead of using MSE, we used $R^2$. When SST is fixed, it can be seen that $R^2$ is inversely proportional to MSE. As a result, for each of the contour plots lighter region corresponds to high performing models and darker regions correspond to poor performing models.

Here are some observations from Figure 1:

i) Row 1 and Row 2 confirm that RFs perform best when the individual trees are deep.

ii) Row 1 and Row 3 show that the best values for MSL is higher for binary response than for continuous responses (BS dataset). This suggests that models for BS data may be prone to over-fitting. However this observation is based on just a single dataset, and we need additional analysis for generalizability.

iii) Row 2 and Row 3 shows performance for Max_p. Optimal values are usually above $\sqrt{p}$ except in PLL data where performance decreases above $\sqrt{p}$. Tuning the HP in some cases might improve performance as is clearly reflected in the Bike Share data and PLL data.

iv) We can see from Figure 1 that the behavior is smooth across our chosen grid. Hence model performance is unlikely to fluctuate greatly with slight changes in some HP configuration.

v) Even though Figure 1 suggests the presence of interactions, the two-dimensional PDPs in Figure 2 show that they are not strong except the interaction in PLL data from Depth vs MSL.

| HP pairs | HL | PLL | BS |
|---|---|---|---|
| Depth vs MSL | 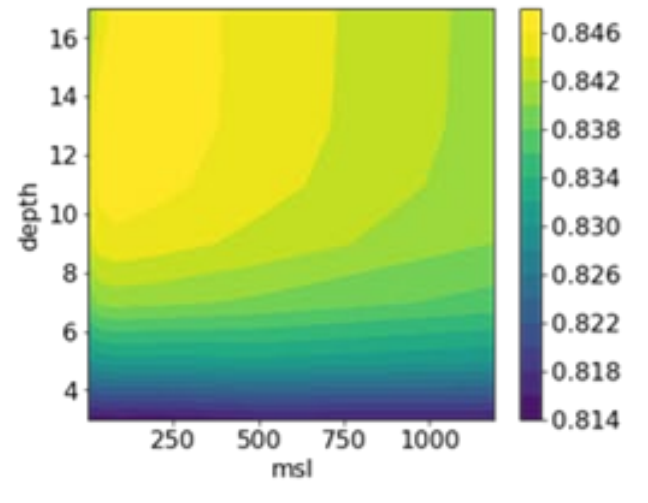 | 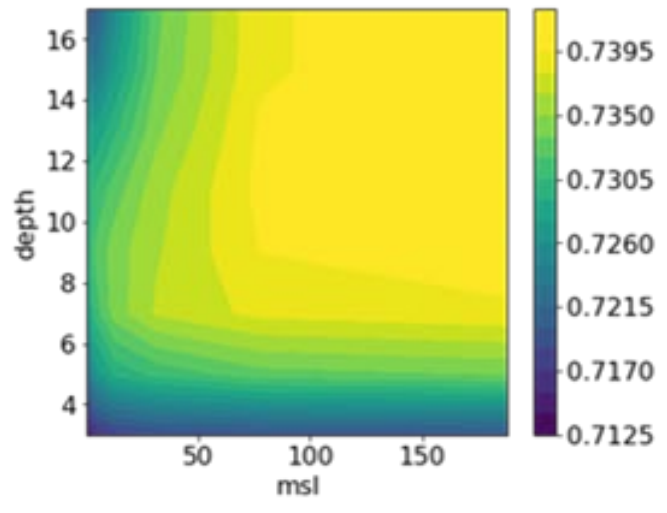 | 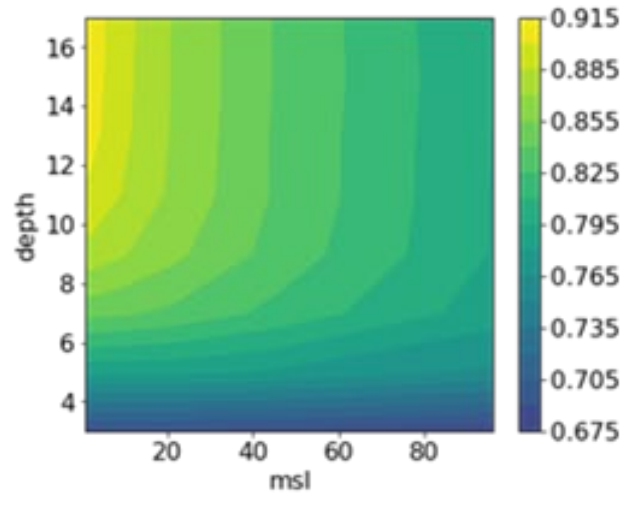 |
| Depth vs Max_p | 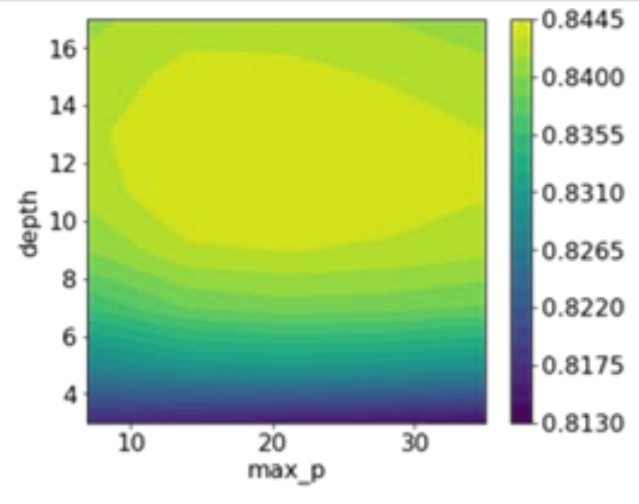 | 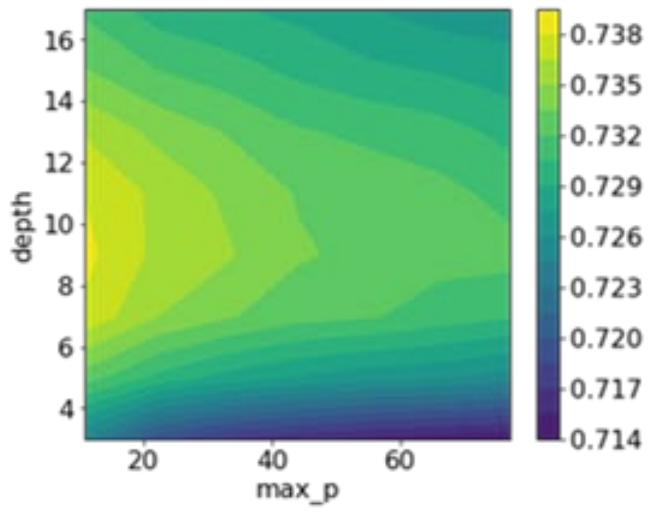 | 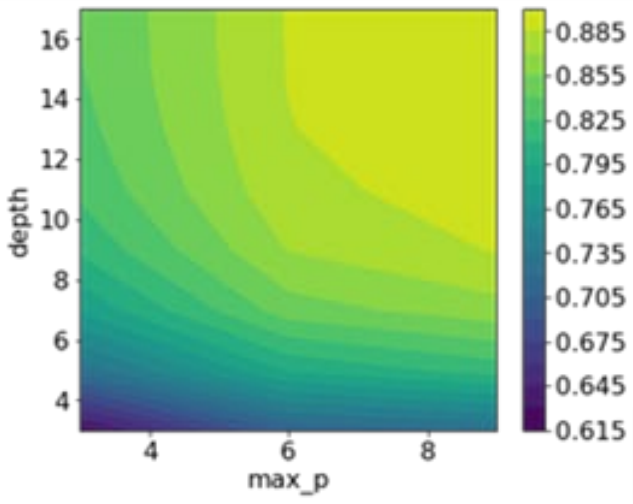 |
| MSL vs Max_p | 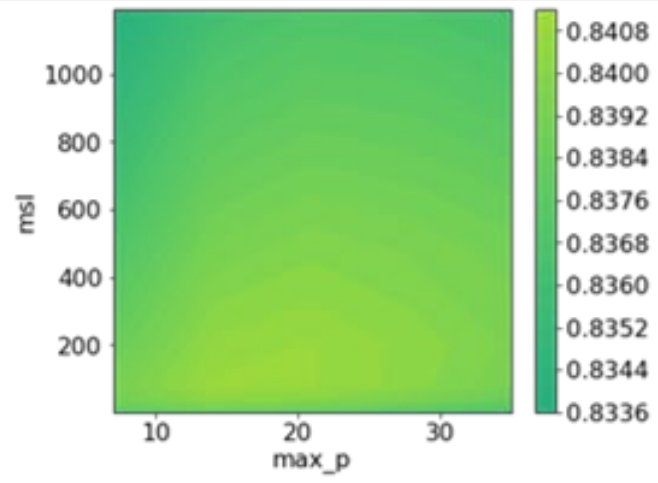 | 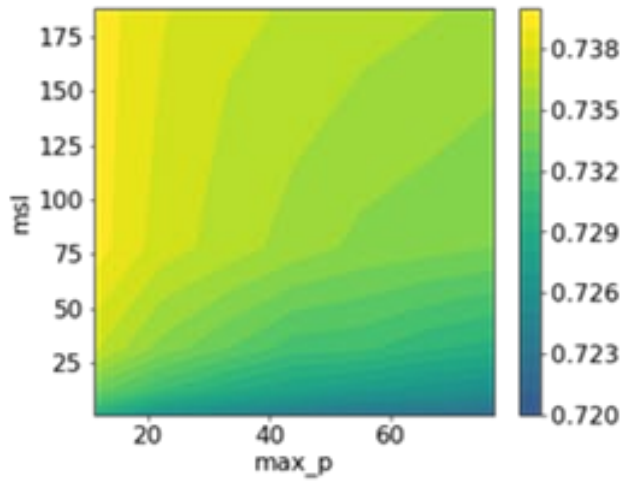 | 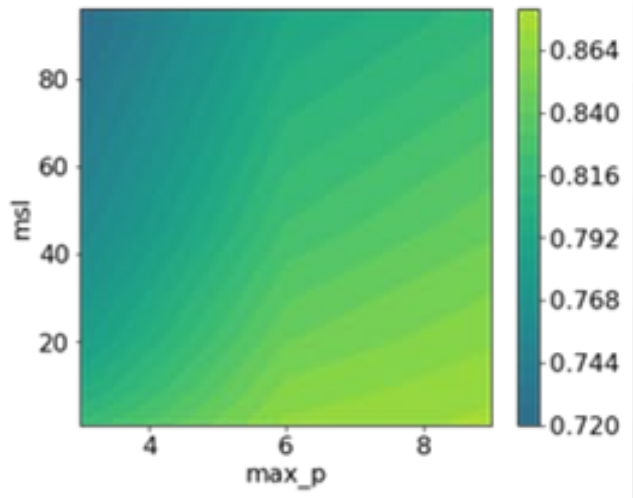 |

Figure 1: Contour plots of selected HP pairs in RF models

| HP pairs | HL | PLL | BS |
|---|---|---|---|
| Depth vs MSL | 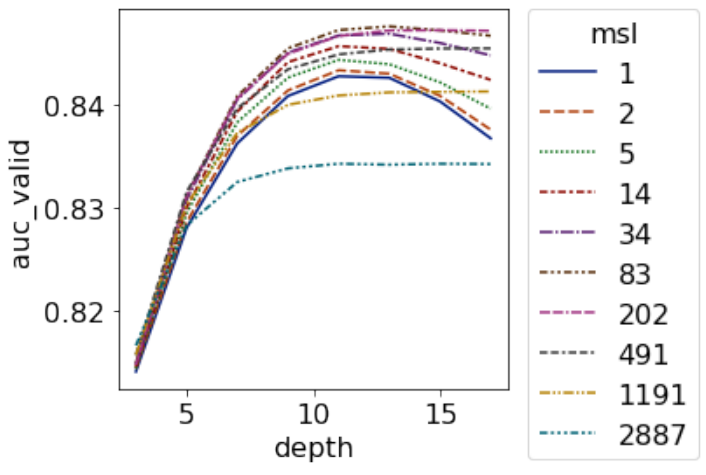 | 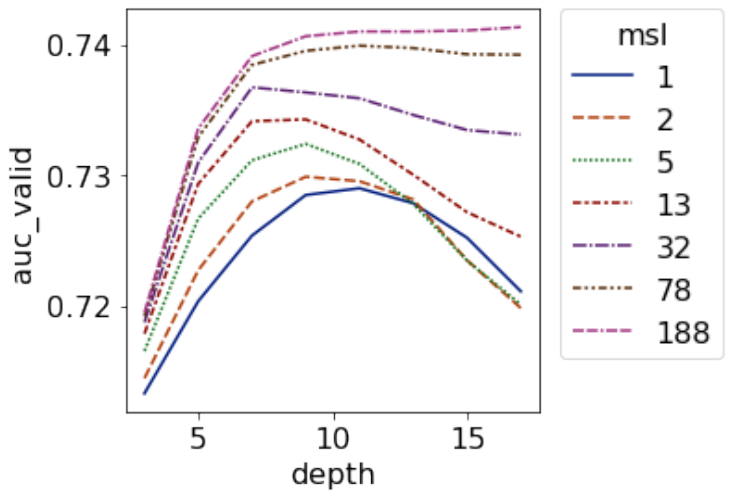 | 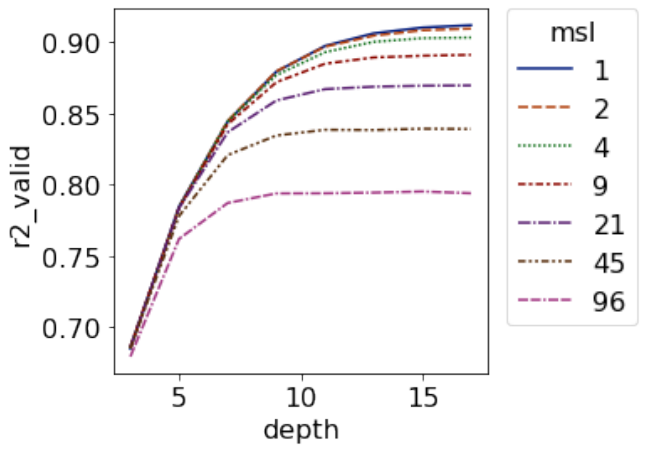 |
| Depth vs Max_p | 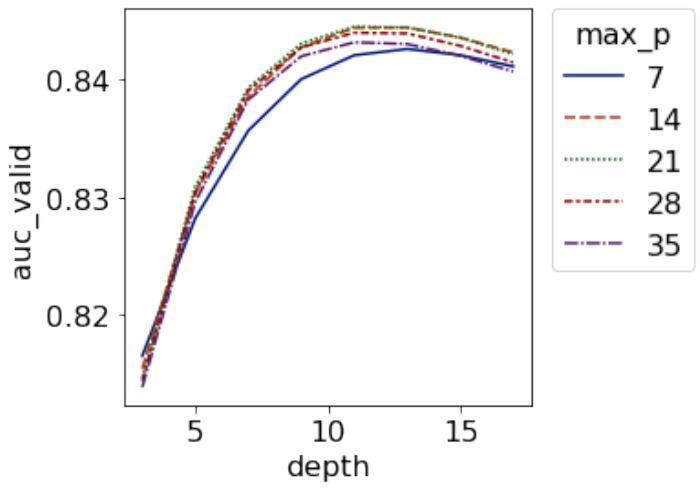 | 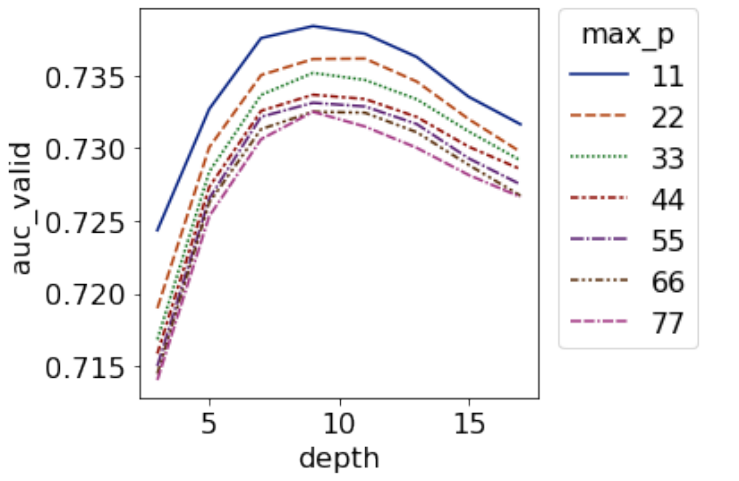 | 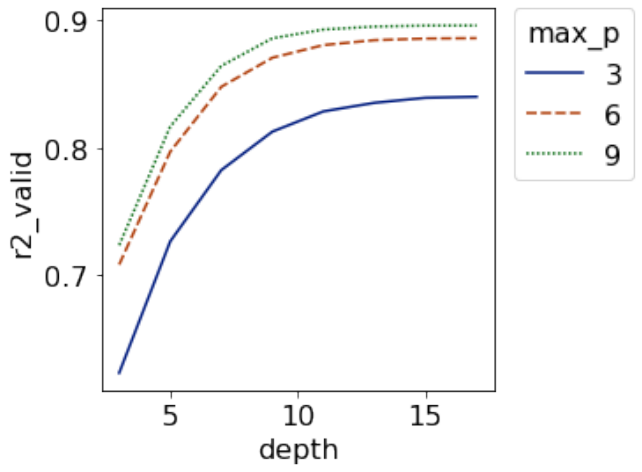 |
| MSL vs Max_p | 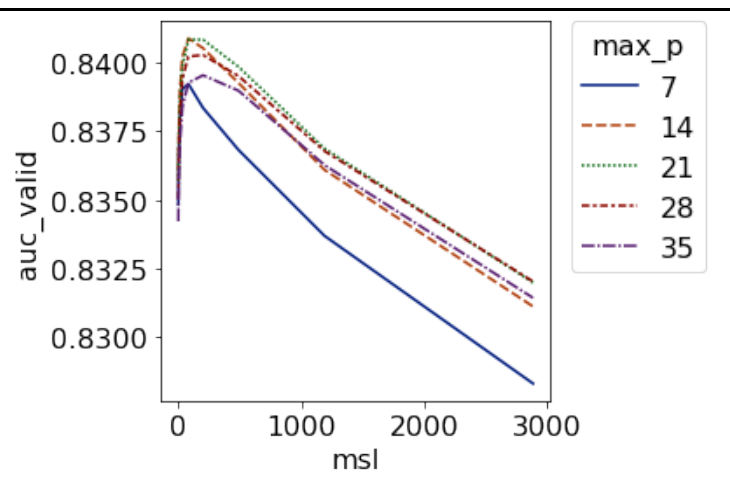 | 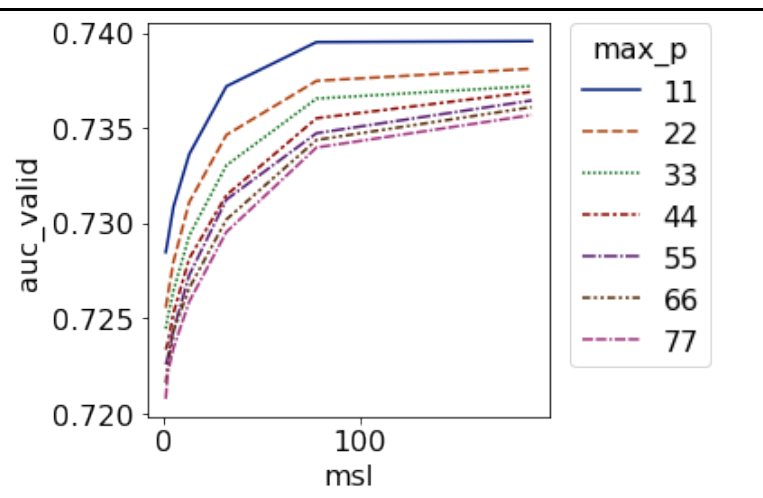 | 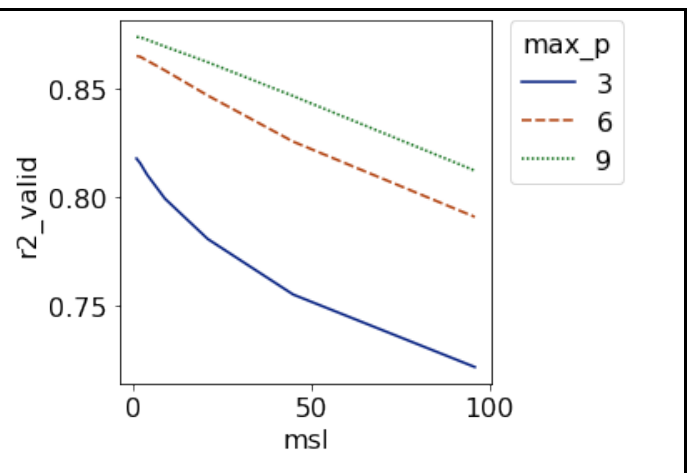 |

Figure 2: Interaction plots of selected HP pairs in RF models

### 4.2. Local Behavior in High-Performing Regions of the Hyper-Parameter Space

We also examined the behavior of the HPs, especially interactions, among the top models in our investigation. This type of interactions is key to possible existence of multiple competing models. To isolate these interactions, we looked at the rank correlation between the HPs in the 50 best performing models in our grid.

The rank correlation matrices in Table 15 show that, in most cases, Max_p has non-negligible correlation with Depth and MSL. This indicates that, apart from tuning the strongest contributing factors Depth and MSL, additional tuning of Max_p may increase performance in an already well-performing model. Due to the interactions between the HPs, we can get several competing models with similar performance but slightly different HP combinations. This can be seen in Figure 1 where there are many models with equally high performance.

Table 15: Rank correlations of HPs in RF in top 50 models

| Data | | |
|---|---|---|
| HL | 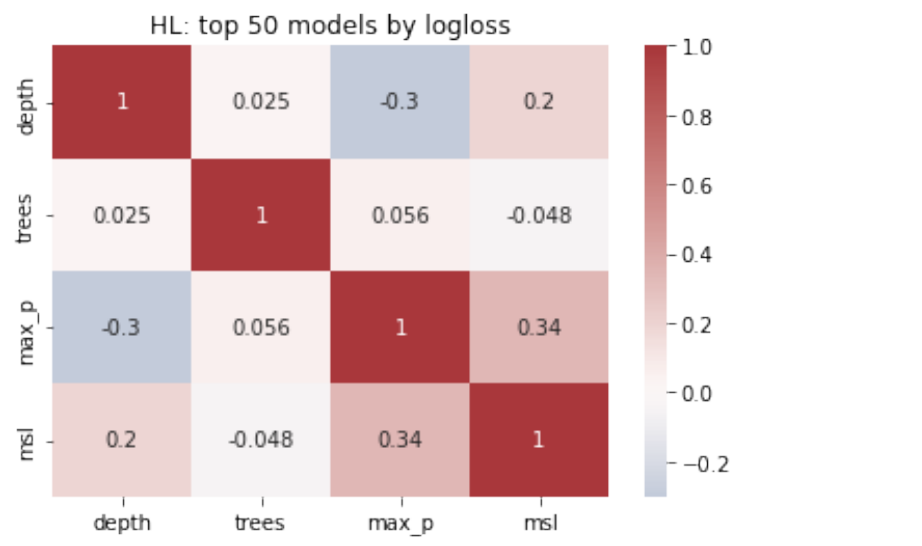 | 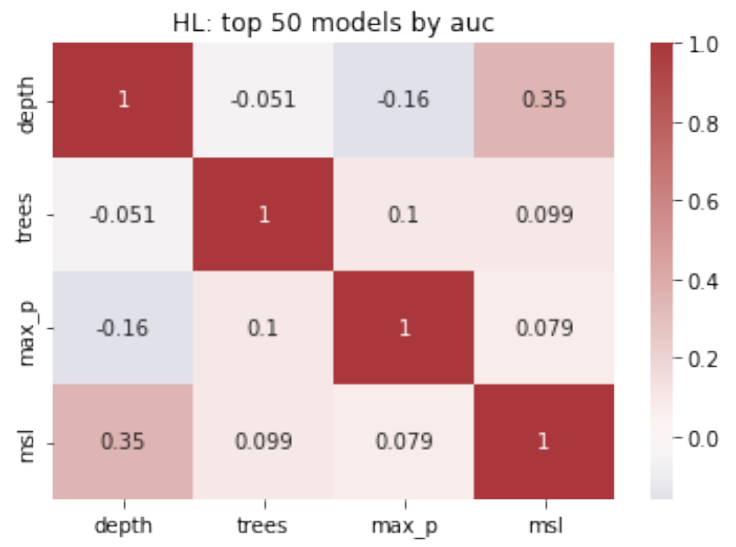 |
| PLL | 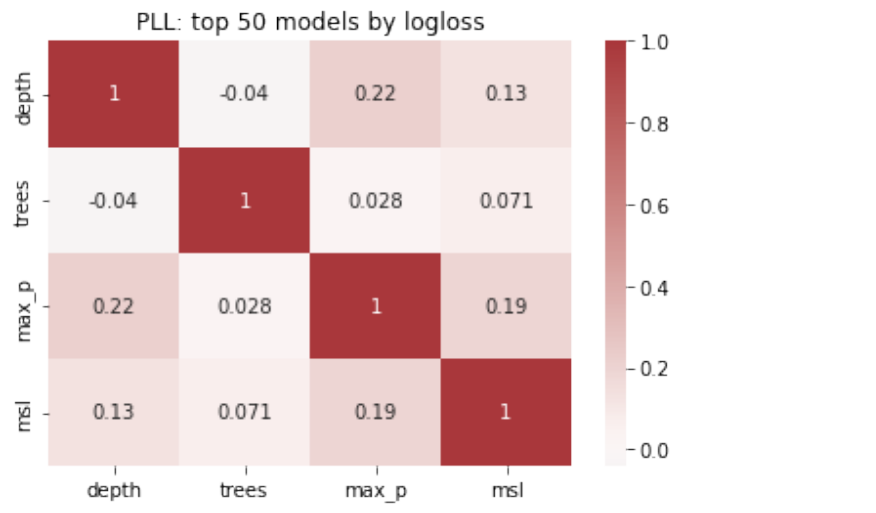 | 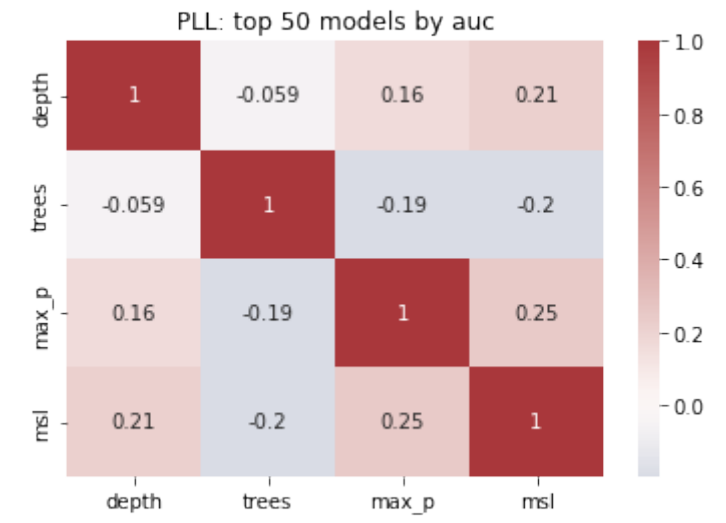 |
| BS | 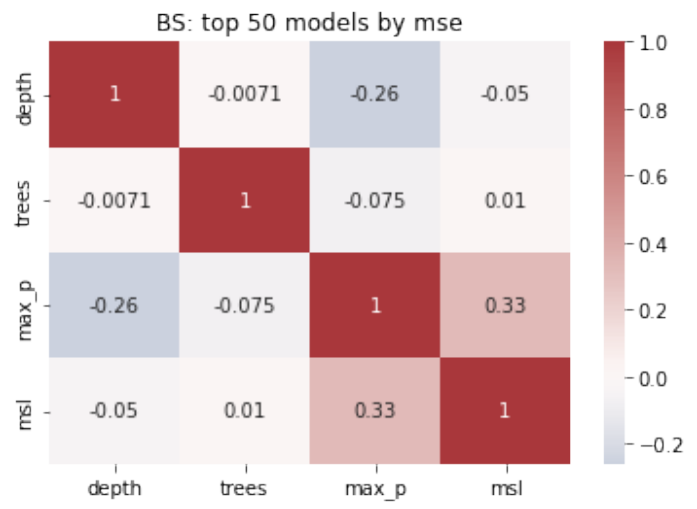 | |

### 4.3. Robustness and Overfitting

To examine the evidence for over-fitting of the best-fitting models, we considered the gap in the metrics evaluated on the training and validation subsets as described in Section 3.1.

Figure 3 shows the gap statistics with training Logloss or MSE on the Y-axis and validation Logloss or MSE on the X axis. The plotting symbols are colored by the levels of the relevant HP. Hence in each plot, the typical region of good models lie on the bottom left corner of the plots. Models that lie on the bottom right are typically under-fit models, and those on the upper half of the plot have high gap statistics and hence likely over-fit.

<table>
<tr><td>Datasets</td><td colspan="2">Depth</td><td colspan="2">MSL</td></tr>
<tr><td>HL</td><td colspan="2">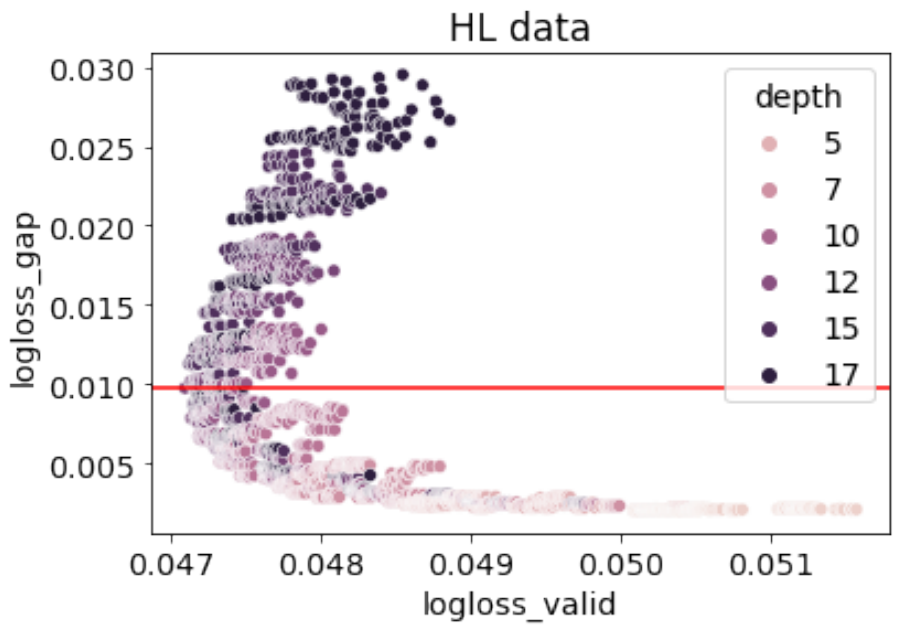
</td><td colspan="2">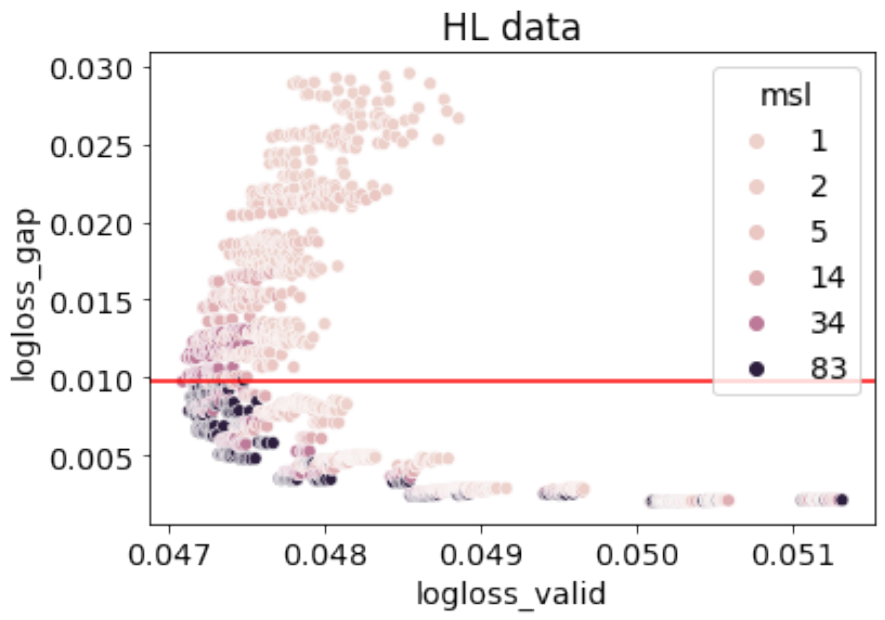
</td></tr>
<tr><td>PLL</td><td colspan="2">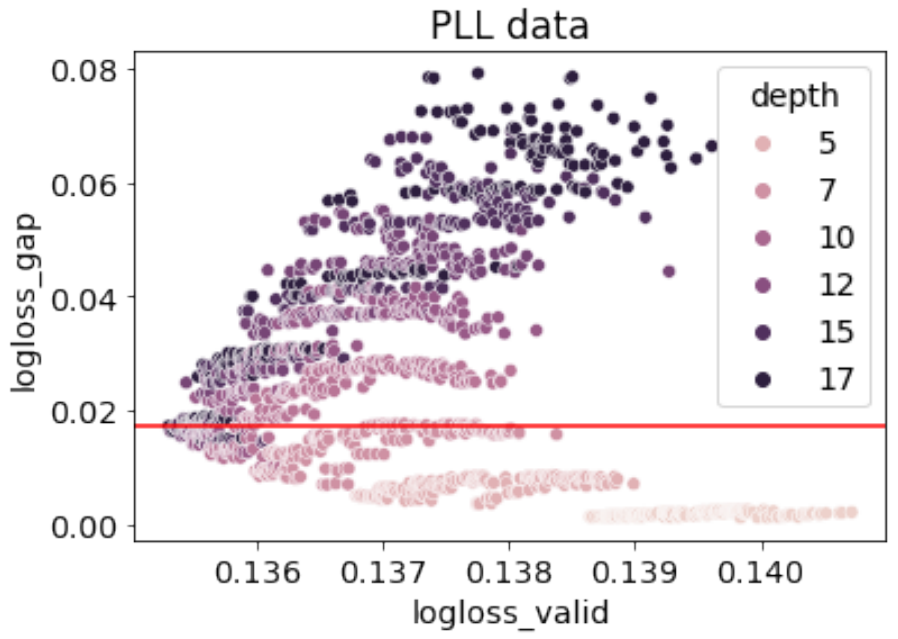
</td><td colspan="2">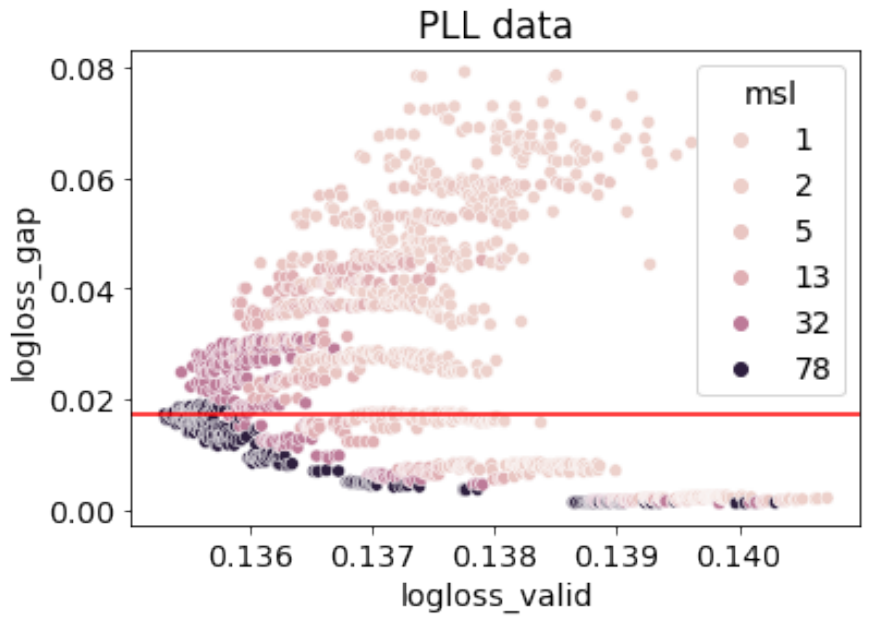
</td></tr>
<tr><td rowspan="2">BS</td><td>Depth</td><td colspan="2">MSL</td><td>Max_p</td></tr>
<tr><td>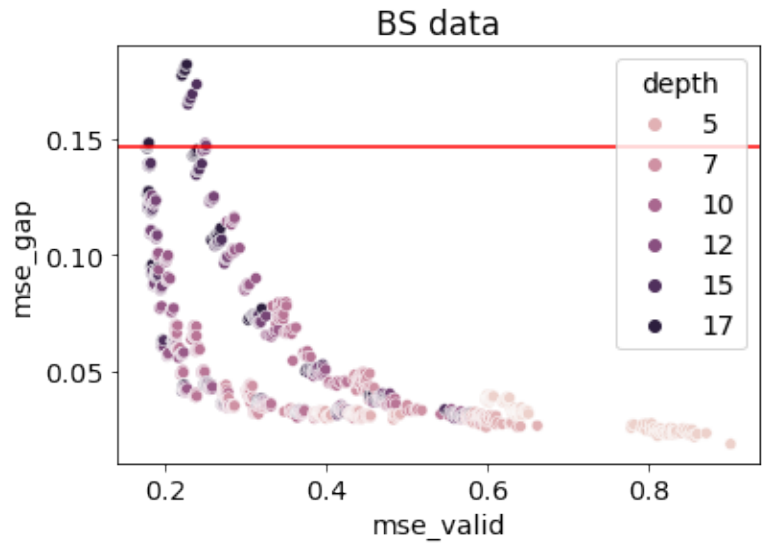
</td><td colspan="2">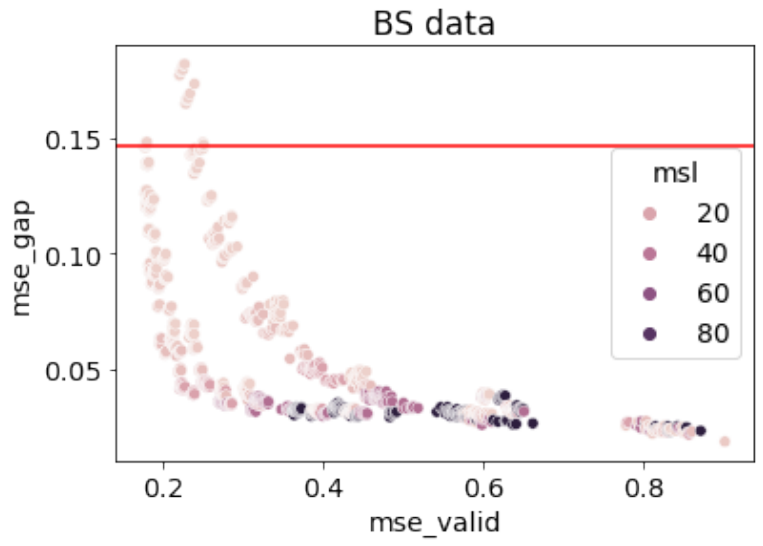
</td><td>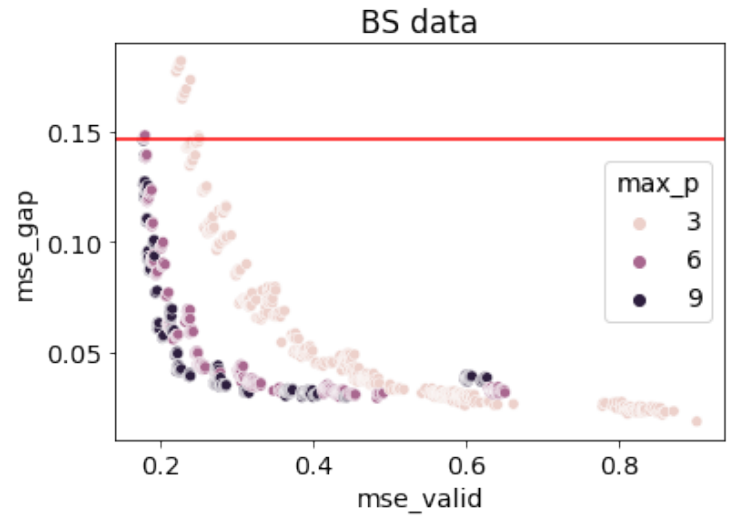
</td></tr>
</table>

Figure 3: Analysis of Gap Metrics to study overfitting:
Red line denotes gap metric of globally best model from grid search

We can observe the following from Figure 3.

i. The panels for MSL show that large values (above a chosen threshold) help control overfitting. However, if MSL is too large, performance in validation set degrades. Hence MSL has to be tuned carefully to obtain optimal and robust RFs.

ii. For the HL and PLL datasets, over-fitted model (the gap values are high) occur mostly when Depth $\geq 15$. Further, as Depth keeps increasing, the Logloss also increases.

iii. The situation is different and more interesting with BS dataset. We observe two separate curves: one consistently below the other. The panel on the right for BS data, with the dot colors corresponding to three levels of Max_p provides a likely explanation. The lower curve has darker dots and corresponds for the most part to higher values of Max_p. When Max_p is 3, the RF models have large gap, and at the extreme, also large MSE_valid. An examination of these models showed they are often unable to pick the more important features or their interactions and hence suffer from large bias in their prediction. Note also that models with large gap statistics do not necessarily have higher MSE_valid.

iv. The relative gap, computed as $\frac{metric_{gap}}{metric_{valid}} \times 100$, of the globally best models for HL, PLL and BS data are 21%, 13% and 83% respectively.

Table 16: Competing RF models for robustness evaluation with BS dataset

| Model | Depth | Trees | Max_p | MSL | MSE_train | MSE_valid | MSE_gap |
|---|---|---|---|---|---|---|---|
| msealt | 15 | 250 | 9 | 9 | 0.13358 | 0.194483 | 0.060903 |
| mseopt | 17 | 400 | 9 | 1 | 0.030322 | 0.17641 | 0.146088 |

This raises a concern whether, for the BS dataset, the globally optimal model in our search is less robust than a competing model with lower gap. To study this further, we compared it with an alternative, well-performing, model for BS dataset in Table 16. Note first that the MSE of the optimal model, resulting from full grid search, is 9% lower than the MSE of the alternate model. Its performance on the training subset is substantially better, and hence its MSE_gap is more than twice that of the alternate model. Note also that the optimal model has a higher Depth and lower MSL compared to the alternate model. So, based on the gap analysis alone, it appears that the alternative model is preferable: slightly worse predictive performance (about 10% higher MSE) but much smaller gap.

However, performance gap does not always translate to lack of robustness, so we studied this further through a robustness analysis using the perturbation methods in (Xu, Vaughan, Chen, Nair, & Sujianto, 2021) **.** This approach involves randomly changing the values of the predictors up to a given budget, using an appropriate metric to measure the new predictive performance in the region, and examining performances for different perturbation budgets. This analysis showed that the globally optimal model and the competing model behaved similarly. Thus, we could not conclude that the model with lower gap is more robust to perturbations.

## 5. Detailed Results in Support of the Findings for XGB models

XGBoost model is a distributed, regularized gradient boosting algorithm that optimizes

$$\sum_{i=1}^{n} L(y_i, h(x_i)) + \sum_{m=1}^{M} \Omega(f_m)\text{, where}$$

$$\Omega(f_m) = \gamma|f_m| + \frac{\lambda}{2}\sum_{j=1}^{|f_m|} w_{mj}^2 + \alpha\sum_{j=1}^{|f_m|} |w_{mj}|\,.$$

Here $|f_m|$ is the number of leaves in the $m-\text{th}$ tree and the L1 ($\alpha$) and L2 ($\lambda$) are penalties on the weights of the leaves in the trees.

We looked at five HPs for the Scikit-learn wrapper interface of XGBoost implementation: Depth, number of Trees, learning rate $\gamma$, L1 penalty on weights of the trees $\alpha$, and L2 penalty on weights of trees $\lambda$. For in depth discussion on each of these parameters, see (Guestrin, 2016) and the package documentation XGBoost API. The idea of boosting stems from fitting weak learners on residuals, hence the individual trees are usually shallow compared to RF. Usual practice is to use shallow trees with a fixed small learning rate (Caruana & Niculescu-Mizil, 2005). For the XGBoost HPs. we used the same grid across all the datasets.

- The learning rate, $\gamma$, was kept low: between 0.01 and 0.2.
- Trees varied from 100 to 500.
- Depth was kept low between 3 and 7.
- The penalty HPs, $\lambda$ and $\alpha$, were geometrically placed from 0 to 10.
- Other parameters are fixed at default values as specified in the XGBoost API.

Table 17: XGBoost HP grid specification (same grid for all datasets)

| HP | Grid specification |
|---|---|
| Learning rate (Lr_rate) | 0.01, 0.03, 0.05, 0.07, 0.1, 0.12, 0.14, 0.16, 0.18, 0.2 |
| Number of estimators (Trees) | 100,200,300,400,500 |
| Depth | 3,4,5,6,7 |
| Alpha - L1 | 0.0, 0.1, 0.46, 2.15, 10 |
| Lambda – L2 | 0.0, 0.1, 0.46, 2.15, 10 |

We ran a complete search over the specified grid and recorded the values of the evaluation metric in both training and validation subset for all HP configurations. The complete grid resulted in a total of 6,250 configurations of HPs.

### 5.1 Global Analysis to Identify Important Hyper-Parameters

In our initial plot based exploratory analysis, we observed that, when the Lr_rate is as low as 0.01, models can perform poorly if not balanced by higher value for Trees (>500). This can drastically impact the results, masking all other significant effects. Hence, in order to tease out the strong effects under reasonable HP configurations, we discarded models with Lr_rate <=0.01. This resulted in a total of 5,625 models.

As noted in Section 3.1, we used ANOVA informally to do a global assessment of the HPs and their interactions by ranking the effect sizes. Table 18-Table 22 report the top 5 effects for each dataset based on both the Logloss and AUC metrics in the case of binary response and based on MSE in the continuous case.

i. In majority of the cases, Depth has the strongest effect followed by Trees and Lr_rate. The magnitudes of the latter two are similar.
ii. Interaction between Trees and Lr_rate affects model performance strongly as expected.
iii. L1 and L2 had the same grid values in our search, however effect of L1 is consistently stronger in ANOVA tables in spite of L1 and l2 having same grid points, suggesting that model performance is more sensitive to change in L1.
iv. Other effects like L2, interaction between Depth and Trees also appear in some cases but they are not in the top 5 effects consistently across the datasets.

Table 18: ANOVA Table for HL Data
Top 5 effects by AUC

| HP | % F | % cum F |
|---|---|---|
| Depth | 40.95 | 40.95 |
| Lr_rate | 15.22 | 56.17 |
| Trees | 10.19 | 66.36 |
| Trees: Lr_rate | 8.28 | 74.64 |
| L1 | 7.91 | 82.55 |

Table 19: ANOVA Table for HL Data
Top 5 effects by Logloss

| HP | % F | % cum F |
|---|---|---|
| Depth | 28.69 | 28.69 |
| Trees: Lr_rate | 22.32 | 51.01 |
| Lr_rate | 18.92 | 69.93 |
| Trees | 14.76 | 84.69 |
| Depth: Trees | 3.79 | 88.48 |

Table 20: ANOVA Table for PLL Data
Top 5effects by AUC

| HP | % F | % cum F |
|---|---|---|
| Trees | 29.87 | 29.87 |
| Lr_rate | 27.75 | 57.62 |
| Depth | 25.63 | 83.25 |
| L1 | 12.39 | 95.64 |
| L2 | 1.93 | 97.57 |

Table 21: ANOVA Table for PLL Data
Top 5 effects by Logloss

| HP | % F | % cum F |
|---|---|---|
| Depth | 35.43 | 35.43 |
| Trees | 23.14 | 58.57 |
| Lr_rate | 16.04 | 74.61 |
| L1 | 11.73 | 86.34 |
| L2 | 3.19 | 89.53 |

Table 22: ANOVA Table for BS Data
Top 5 effects by MSE

| HP | % F | % cum F |
|---|---|---|
| Depth | 40.92 | 40.92 |
| Trees | 29.36 | 70.28 |
| Lr_rate | 18.81 | 89.09 |
| L1 | 3.17 | 92.26 |
| Trees: Lr_rate | 2.76 | 95.02 |

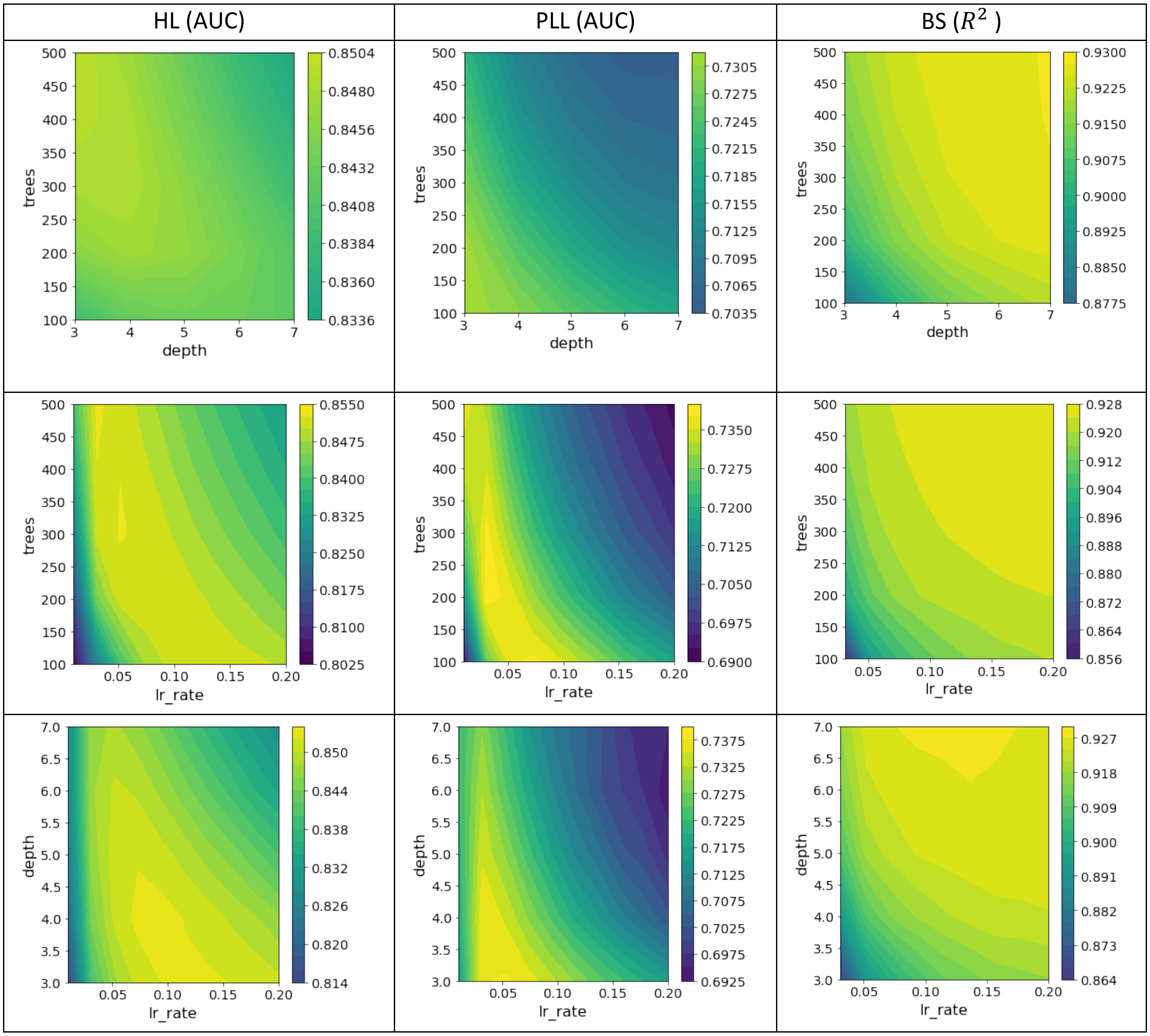

Figure 4: Contour plots for important interaction effects in XGBoost models

In Table 18-Table 22, we observed varying levels of interactions between Trees and Lr_rate, and in one case, Depth and Trees. To understand the nature of this interaction, let us look at the contour plots of the performance for these three pairs of HPs, after averaging across the other HP combinations. For each data set the contour plots are mapped to the same color scale across pairs of HP for ease of comparison. In case of BS data, instead of using MSE, we used $R^2$ which is inversely proportional to MSE for constant TSS (Total Sum of Squares). As a result, for each of the contour plots lighter region corresponds to high performing models and darker regions correspond to poor performing models. The results are shown in Figure 4. We observe the following:

i. The contour plots show the nature of the interactions more clearly. In general, the behavior is smooth across our chosen grid. Hence model performance is stable and changes smoothly with change in HP configuration.
ii. For binary response (HL and PLL), as expected, we observe an inverse relationship between each pair of HPs: if Depth is low, we need higher number of Trees to eliminate bias; if Lr_rate is low, we need higher Depth or higher number of Trees to get good performing models.
iii. For continuous response (BS) good models prefer high Depth, high number of Trees, and high Lr_rate). Hence the may be prone to over-fitting issues. We study this in more detail in Section 5.3. But we do we see the inverse relationship partially in the contour plots for BS data. It is likely that we need a larger grid in this case for the inverse relationship to be viewed clearly.

### 5.2. Local Behavior in High-Performing Regions of the Hyper-Parameter Space

We now restrict attention to the top models in our search and examine presence of additional interactions between HPs. To isolate the interactions, we look at the rank correlation between the HPs in the 50 best performing models in our grid in Table 23. We make the following observations:

Table 23: Rank correlations between HPs in top 50 XGBoost models

<table>
<tr><td>Data</td><td colspan="2"></td></tr>
<tr><td>HL</td><td>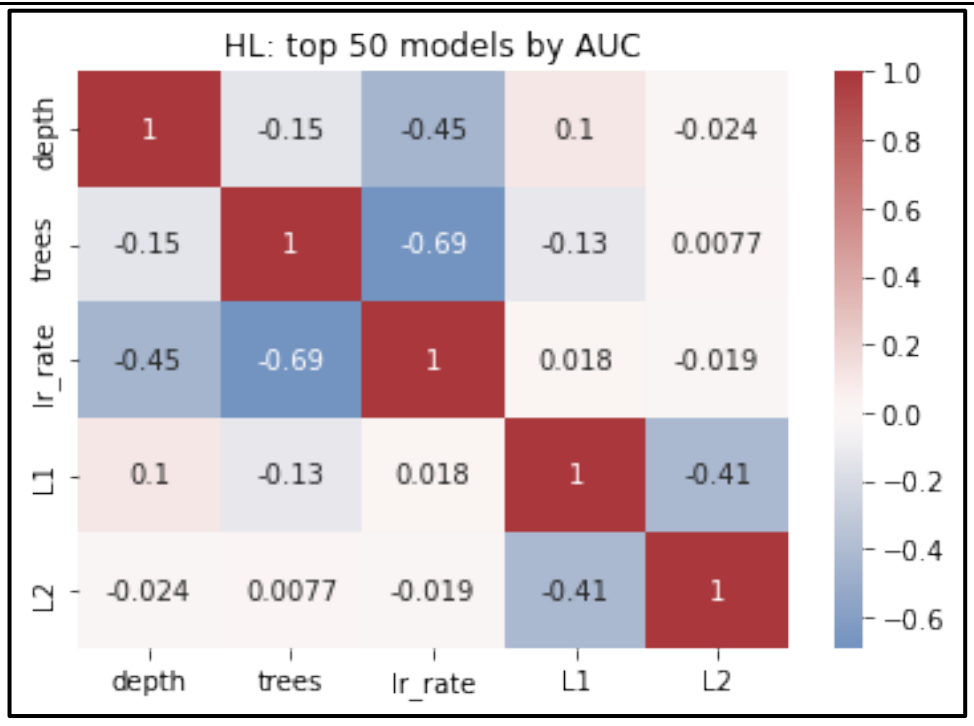
</td><td>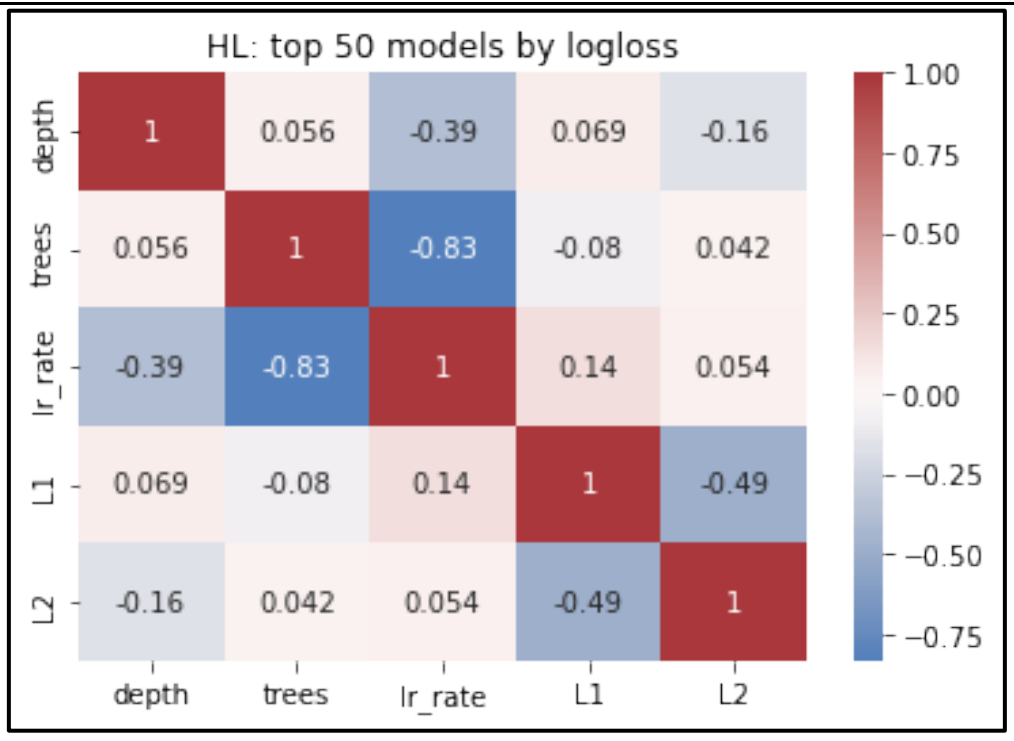
</td></tr>
<tr><td>PLL</td><td>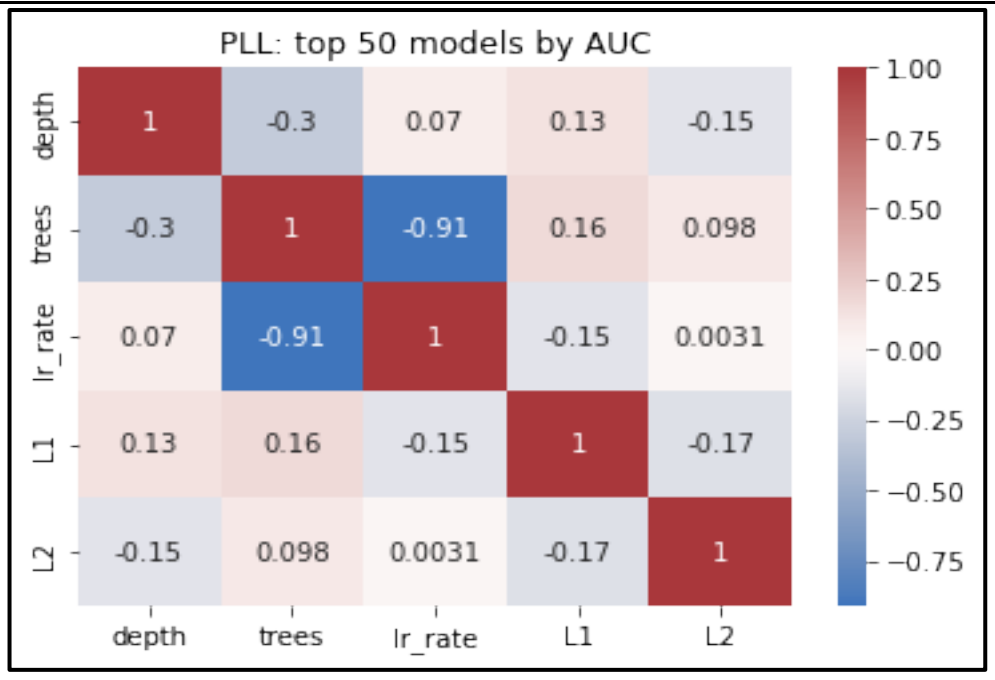
</td><td>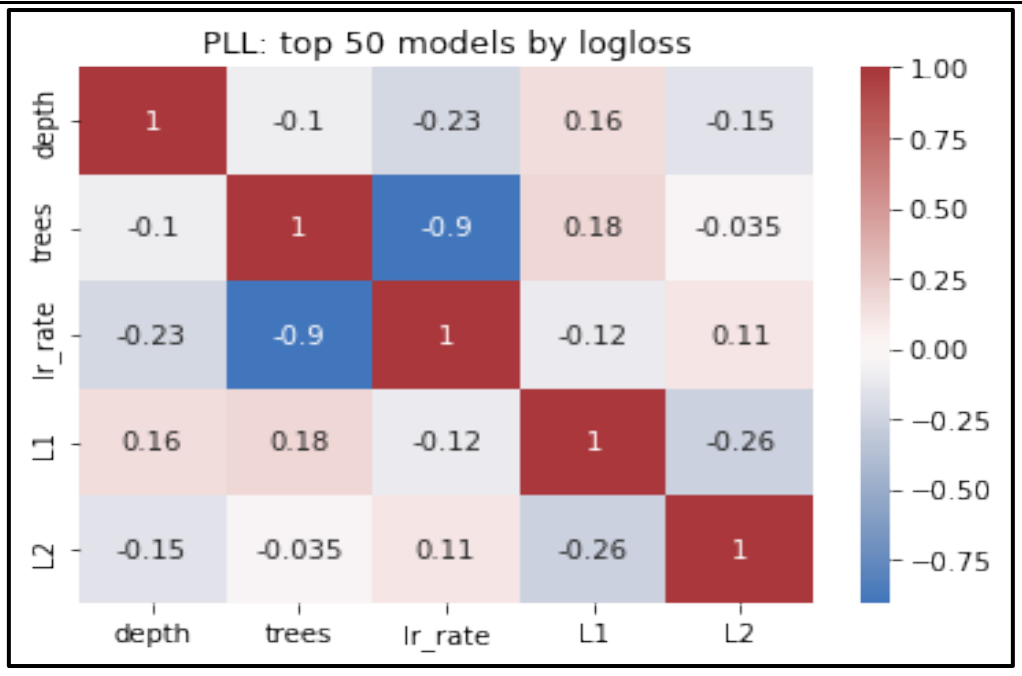
</td></tr>
</table>

| | | |
|---|---|---|
| BS | | 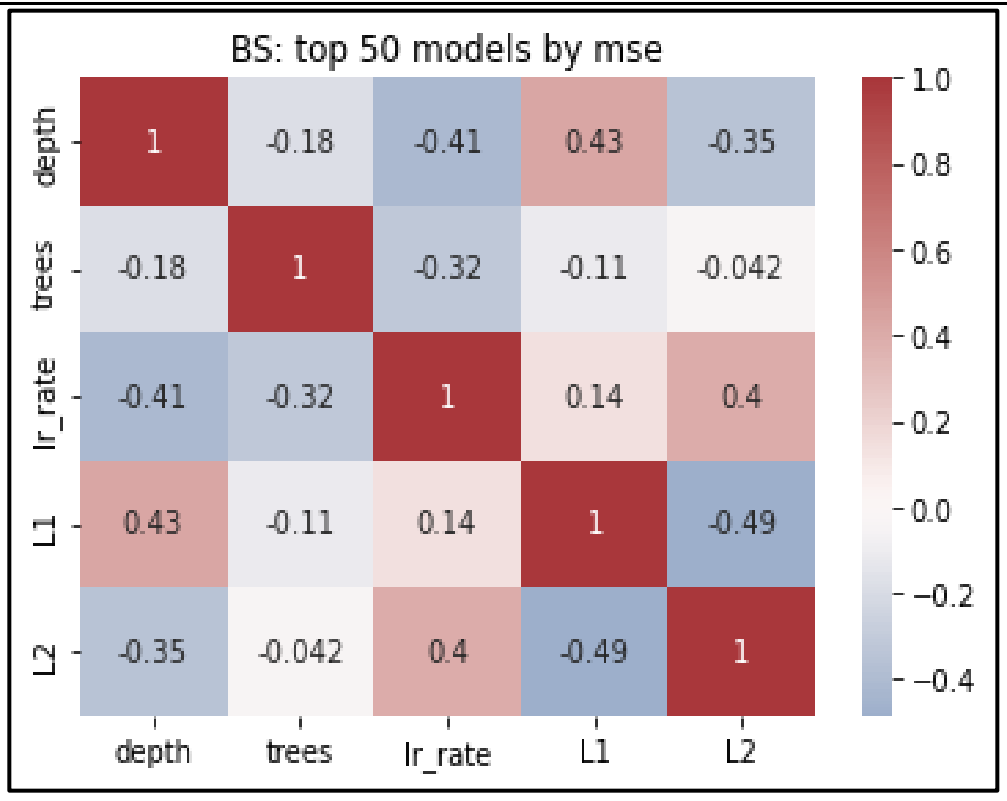  |

i. The top rank correlations are those between Lr_rate and Trees, Depth and Trees, and between Depth and Lr_rate. It turns out that all of these appeared as interactions in the top 10 effects in our ANOVA although we reported only the top 5 effects.

ii. Additionally, we see a significant negative rank correlation between L1 and L2 penalties in the top 50 models specifically for HL and BS data. This suggests that the models usually prefer one form of penalization but not both. We explore this further in the heatmaps of L1 vs L2 for the top 50 models in Table 24 colored by the concentration of models for given configuration (Figure 5-Figure 6) indicate that HL data prefers higher L1 penalty. PLL data prefers both forms of penalization simultaneously (Figure 7-Figure 8) and hence the rank correlation was not as strong between L1 and L2 as in other datasets. The BS data generally prefers L2 regularization (Figure 9).

Table 24: Proportion of top 50 XGBoost models at each L1 and L2 configuration

| | |
|---|---|
| 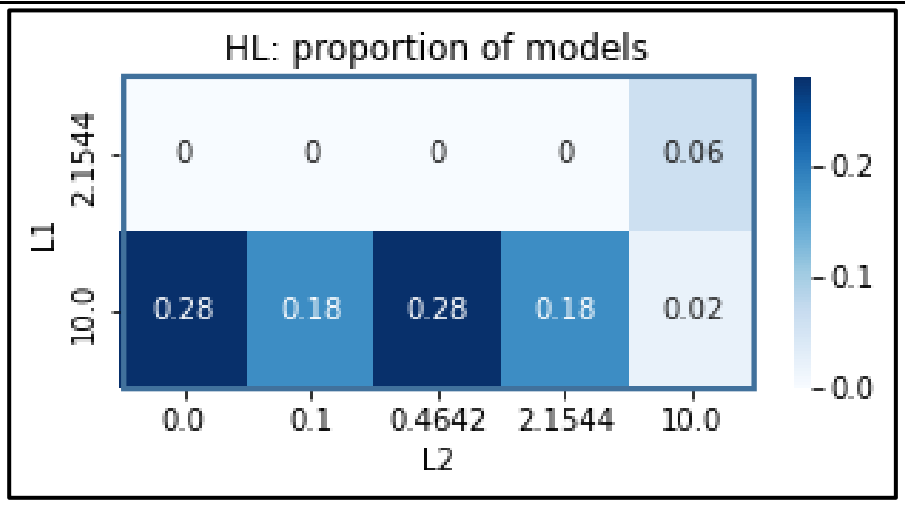  Figure 5: HL Data – Concentration of Top 50 models by AUC | 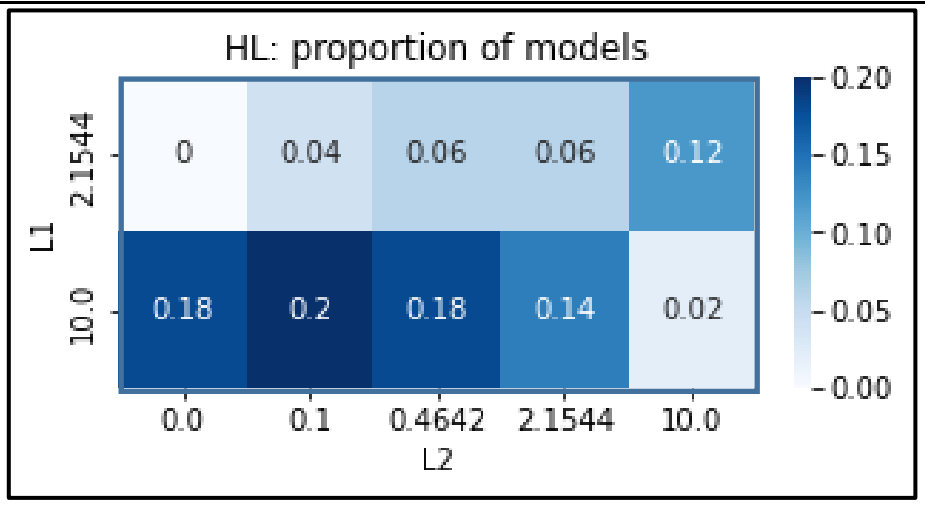  Figure 6: HL Data – Concentration of Top 50 models by Logloss |

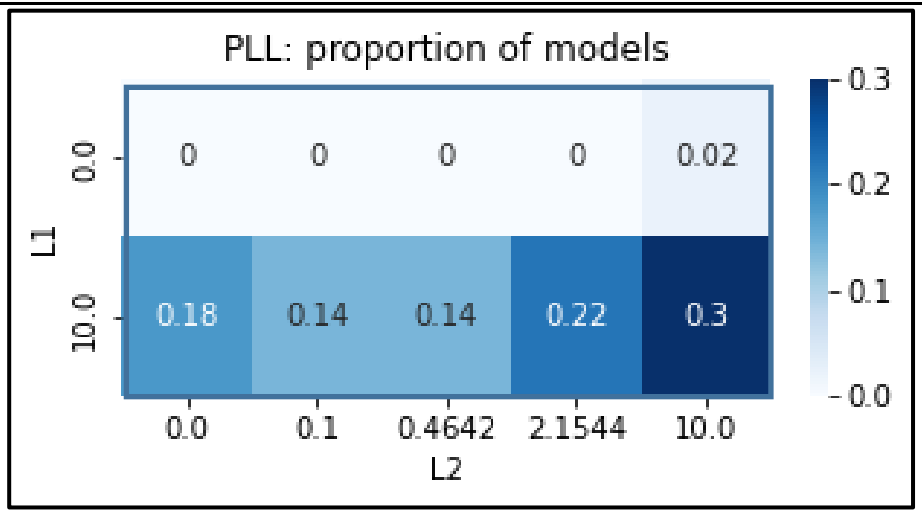

Figure 7: PLL Data – Concentration of Top 50 models by AUC

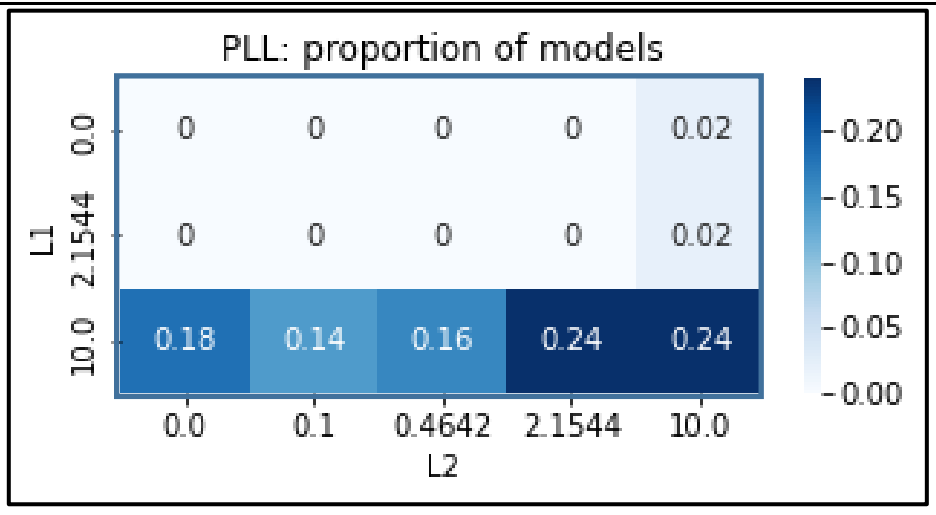

Figure 8: PLL Data – Concentration of Top 50 models by Logloss

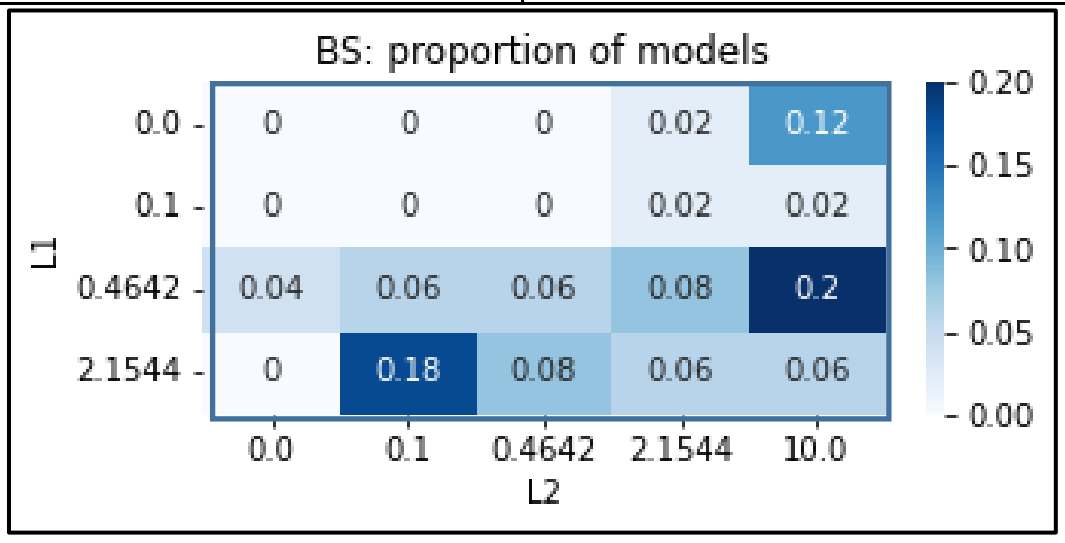

Figure 9: BS Data – Concentration of Top 50 models by MSE

iii. In the BS data, we additionally observe that a high Depth is positively associated with high L1 to counteract the overfitting. However Depth is negatively correlated with L2 which is counterintuitive. This occurs due to the strong negative correlation between L1 and L2 which in turn drives the negative correlation between Depth and L2.

iv. There also exists a positive correlation between Lr_rate and the L1 and L2 penalty which is once again explained by the high regularization required to control over-fitting caused by a high Lr_rate.

These minor interactions (L1 vs L2, Lr_rate vs L1/L2, Depth vs L1/L2, etc.), which did not appear to be strong in the global ANOVA table, can still contribute to the existence of multiple competing models in the HP space.

### 5.3. Robustness and Overfitting

XGBoost models can over-fit due to a high Lr_rate, large value of Trees, or large value of Depth. We looked at the gap in the metrics as described in Section 3.1. Figure 10 is a plot of the Logloss/MSE gap statistics on the Y-axis and the validation Logloss/MSE on the X axis and color the plots by the levels of each HP. Hence in plot the typical region of good models lie on the bottom left corner of the plots. Models that lie on the bottom right are typically under-fit models and models that lie on the upper half of the plot are models with high gap statistics and hence models that have likely over-fit. We have also marked the gap of the optimal model in our search with the red horizontal line.

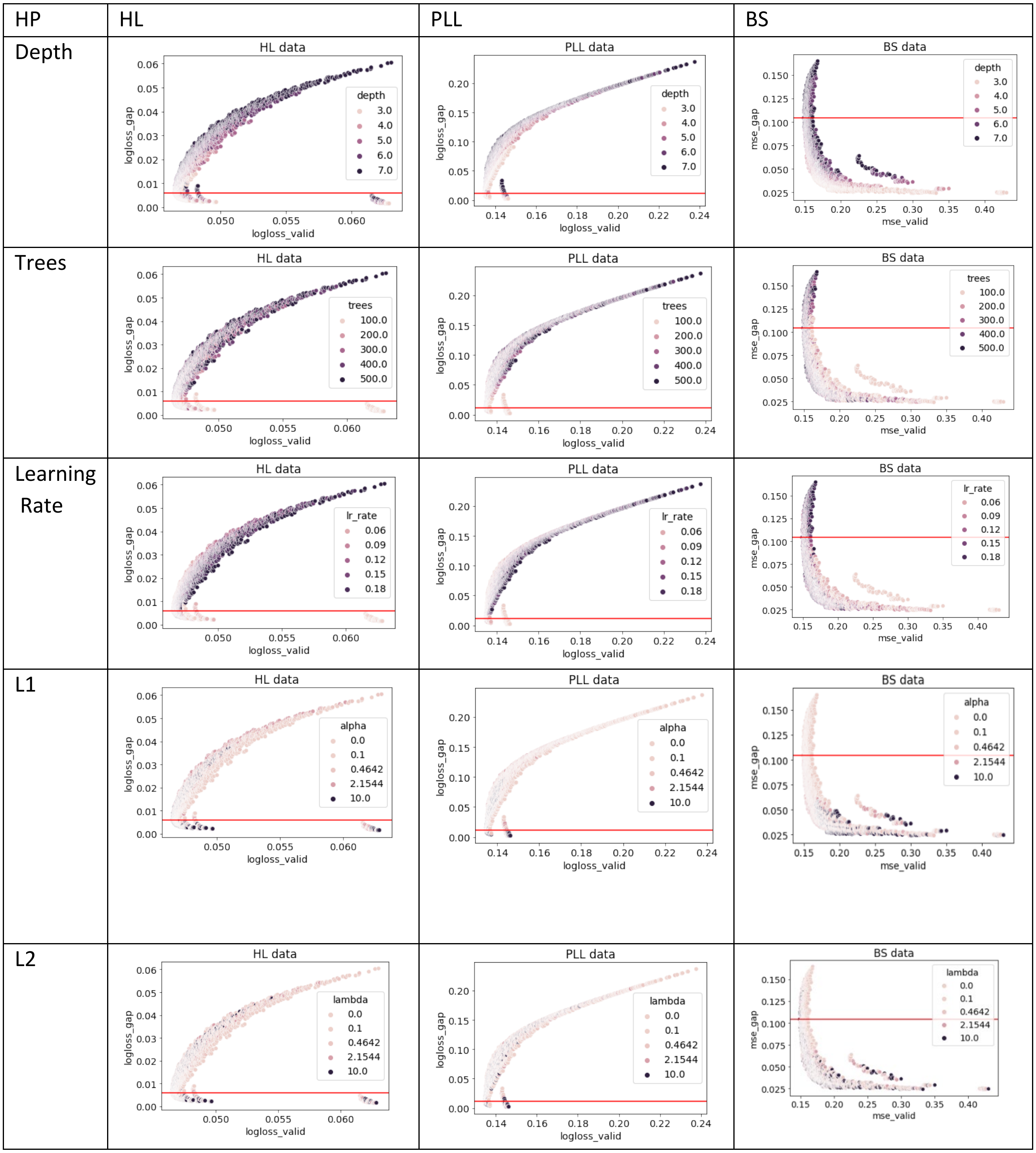

Figure 10: Over-fitting in XGBoost models (red line marks gap statistic of the optimal model in our grid search)

We make the following observations from Figure 10:

i. The models that over-fit usually have high values for more than one of the following HPs: Depth, Lr_rate and Trees. These models either have low regularization penalties or they are models which require more regularization than currently allowed in our grid.

ii. The optimal model in case of BS data has a much larger gap value than those for HL and PLL data. The relative gaps for HL, PLL and BS data are 14%, 8% and 70% respectively. Thus, the best HP configuration for the BS data maybe over-fitting and hence less robust than the alternate model with lower training and validation gap in Table 25.

Table 25: Competing XGB models for robustness evaluation in BS data

| Model | Depth | Trees | Lr_rate | alpha | lambda | MSE_train | MSE_valid | MSE_gap |
|---|---|---|---|---|---|---|---|---|
| **opt** | 6.0 | 400 | 0.1789 | 0.0 | 10.0 | 0.0446 | 0.1487 | 0.1041 |
| **alt** | 3.0 | 200 | 0.0522 | 0 | 10.0 | 0.2285 | 0.2540 | 0.0255 |

As with the case for RF models, we used the toolbox in (Xu, Vaughan, Chen, Nair, & Sujianto, 2021) to do a robustness analysis by perturbing the predictors. This showed that the best-fitting model was less robust in terms of this particular analysis. While the conclusion here is different from the one for RF, the key takeaway is that the top model resulting from HPO search may be less robust (over-fitting or other issues), especially in case of continuous responses. Thus, other considerations such as robustness should be considered in picking the best HP configuration.

As a side note, we investigated the usefulness of two additional parameters for regularization and reducing overfitting. These were: Min_child_weight (minimum value for the total weight in a leaf node) and colsample_bytree (sub-sampling ratio of columns). See XGBoost documentation (XGBoost Parameters, 2022) for more details. We chose two models, the globally optimal one and another model that had serious over-fitting due to lack of regularization. We performed additional tuning on these models using these two parameters. There was not much gain for the optimal model. However, there was indeed a large change for the over-fit model, and the gap statistic lowered considerably. So it appears that these two additional HPs can be used as alternate forms of regularization, but any broad conclusion needs further study.

## 6. Detailed Results in Support of the Findings for FFNN models

We looked at seven HPs for the Keras FFNN models with one to two hidden layers. The parameters of interest are the sizes of Layer1 and Layer2 (if a second layer is present), three forms of regularization (Dropout, L1 and L2), Lr_rate of the optimizer Adam, and the batch size in the mini batch shuffling that we deployed. We determined the grid for each dataset after some small-scale explorations to rule out extremely poor configurations. Hence the grids differed slightly across datasets, and this was especially so for the regularization HPs.

- The size of Layer1 was tuned from 32 to 512 nodes. Layer2 was either absent or tuned between 16 and 64 nodes.
- Dropout works by randomly disabling neurons and their corresponding connections. This prevents the network from relying too much on single neurons and forces all neurons to learn to generalize better. Depending on the dataset, Dropout was varied from 10% to 50%.
- L1 and L2 are regularization on the kernel weights of each layer, but the bias is left un-regularized. Initial exploration showed that the models were highly sensitive to L1 penalty, and hence it was kept less than 0.001. The L2 was allowed to be slightly higher.

- A large batch size can make the optimizer slow whereas extremely small batch can make it unstable. Extremely small batch sizes should be avoided especially in case of imbalanced data. We used 1,000 epochs and early stopping to train the model. Batch_size was also data dependent, and we varied the batch size to cover between 5 to 20% of the data.
- Finally, Lr_rate was kept low and varied from 0.0005 to 0.003 (default Lr_rate in Adam optimizer is usually 0.001).
- Other HPs are set at default values. For more details on HP tuning for FFNNs, see (TF Keras API, 2015)

Table 26: Grid specification for FFNN models

| HP | HL | PLL | BS | Sim |
|---|---|---|---|---|
| Lr_rate | 0.0005, 0.0007, 0.001, 0.003 | 0.0005, 0.0007, 0.001, 0.003 | 0.0005, 0.0007, 0.001, 0.003 | 0.0005, 0.0007, 0.001, 0.003 |
| Batch_size | 10K, 20K, 30K | 2K, 4K, 6K, 8K | 500, 1000, 1500, 2000 | 1500, 3000, 4500, 6000 |
| Layer1 | 64, 128, 256, 512 | 64, 128, 256, 512 | 64, 128, 256, 512 | 64, 128, 256, 512 |
| Layer2 | 0, 16, 32 | 0, 16, 32 | 0, 32, 64 | 0, 16, 32 |
| L1 | 0, .0005, .001,.005 | 0, .0005, .001,.01 | 0, 0.0005, .001 | 0, .0005, .001,.005 |
| L2 | 0, .005, .01 | 0, .005, .01 | 0, 0.005, .01 | 0, .005, .01 |
| Dropout | 0, 0.3, 0.5 | 0, 0.3, 0.5 | 0, 0.1, 0.2, 0.3 | 0.1, 0.2, 0.3 |

We ran a complete search over the specified grid and recorded the values of the evaluation metric in both training and validation subset for all HP configurations. The complete grid results in a total of 5184 configurations of HP for the HL data and 6912 configurations of HPs for the other three datasets.

## 6.1 Global Analysis to Identify Important Hyper-Parameters

As noted in Section 3.1, we used ANOVA informally to do a global assessment of the HPs and their interactions by ranking the effect sizes. Table 27-Table 32 reports the top 5 effects for each dataset based on both Logloss and AUC in case of binary response and based on MSE in the continuous case. We make the following observations.

i. Dropout, L1 and Layer2 are usually the top ranking HPs, with their relative importance varying with the datasets. For example, Layer2 did not appear in the top 5 effects for PLL data and was included in the top 5 effects for the SIM data only through interaction with L1. L1 appeared in the top 5 effects in all datasets except for HL data.

ii. Lr_rate ranked higher with continuous responses (BS and SIM datasets) than with binary responses. However, in all cases, the total contributions of the F-statistics were less than 10%. We looked at the boxplots (Figure 11**Error! Reference source not found.**) to study the marginal distribution of the metrics with respect to the Lr_rate and observed that there was no discernable difference between the plots from the binary and continuous response. Additionally, it was observed that in general as Lr_rate increased average model performance deteriorated. The default value for Lr_rate in ADAM optimizers is 0.001, hence it is advisable to tune Lr_rate with upper boundary set at 0.001.

iii. Batch_size did not appear in the top 10 effects in any of the tables, though here we are only reporting top 5 effects. This suggests that it does not have a strong direct effect on model performance as long as it is kept within 5 to 20% of the number of observations in the training subset.

iv. Most of the significant interaction effects are that between Layer2 and some regularization HP or between the regularization HPs themselves, but there is no interaction that is consistently most important across the tables. Notable among the interactions are between
   - sizes of Layers and the regularization HPs – for example, Layer2 vs dropout in HL data and L1 vs Layer2 in SIM data,
   - within the regularization HPs – for example, L1 vs dropout and L1 vs L2 in PLL data.

Table 27: ANOVA Table for HL data
Top 5 effects ranked by AUC

| HP | % F | % cum F |
|---|---|---|
| Dropout | 44.00 | 44.00 |
| Layer2 | 24.60 | 68.60 |
| Layer2:Dropout | 21.56 | 90.16 |
| Layer1 | 1.75 | 91.91 |
| Layer1:Dropout | 0.90 | 92.81 |

Table 28: ANOVA Table for HL data
Top 5 effects ranked by Logloss

| HP | % F | % cum F |
|---|---|---|
| Dropout | 44.26 | 44.26 |
| Layer2 | 27.55 | 71.81 |
| Layer2:Dropout | 13.27 | 85.08 |
| Layer1 | 2.21 | 87.29 |
| Lr_rate | 1.74 | 89.03 |

Table 29: ANOVA Table for PLL data
Top 5 effects ranked by AUC

| HP | % F | % cum F |
|---|---|---|
| L1 | 41.77 | 41.77 |
| L1:Dropout | 14.91 | 56.68 |
| L1:L2 | 8.39 | 65.07 |
| Dropout | 6.99 | 72.06 |
| L2 | 6.36 | 78.42 |

Table 30: ANOVA Table for PLL data
Top 5 effects by Logloss

| HP | % F | % cum F |
|---|---|---|
| L1 | 26.56 | 26.56 |
| L2 | 25.03 | 51.59 |
| L1:L2 | 25.01 | 76.60 |
| L1:Dropout | 3.28 | 79.88 |
| Dropout | 2.41 | 82.29 |

Table 31: ANOVA Table for BS data
Top 5 effects ranked by MSE

| HP | % F | % cum F |
|---|---|---|
| Dropout | 47.97 | 47.97 |
| Layer2 | 27.78 | 75.75 |
| L1 | 5.93 | 81.68 |
| Lr_rate | 4.15 | 85.83 |
| Layer1 | 2.18 | 88.01 |

Table 32: ANOVA Table for SIM data
Top 5 effects ranked by MSE

| HP | % F | % cum F |
|---|---|---|
| L1 | 64.09 | 64.09 |
| Dropout | 9.7 | 73.79 |
| Lr_rate | 8.66 | 82.45 |
| Lr_rate:Layer1 | 2.88 | 85.33 |
| L1:Layer2 | 2.34 | 87.67 |

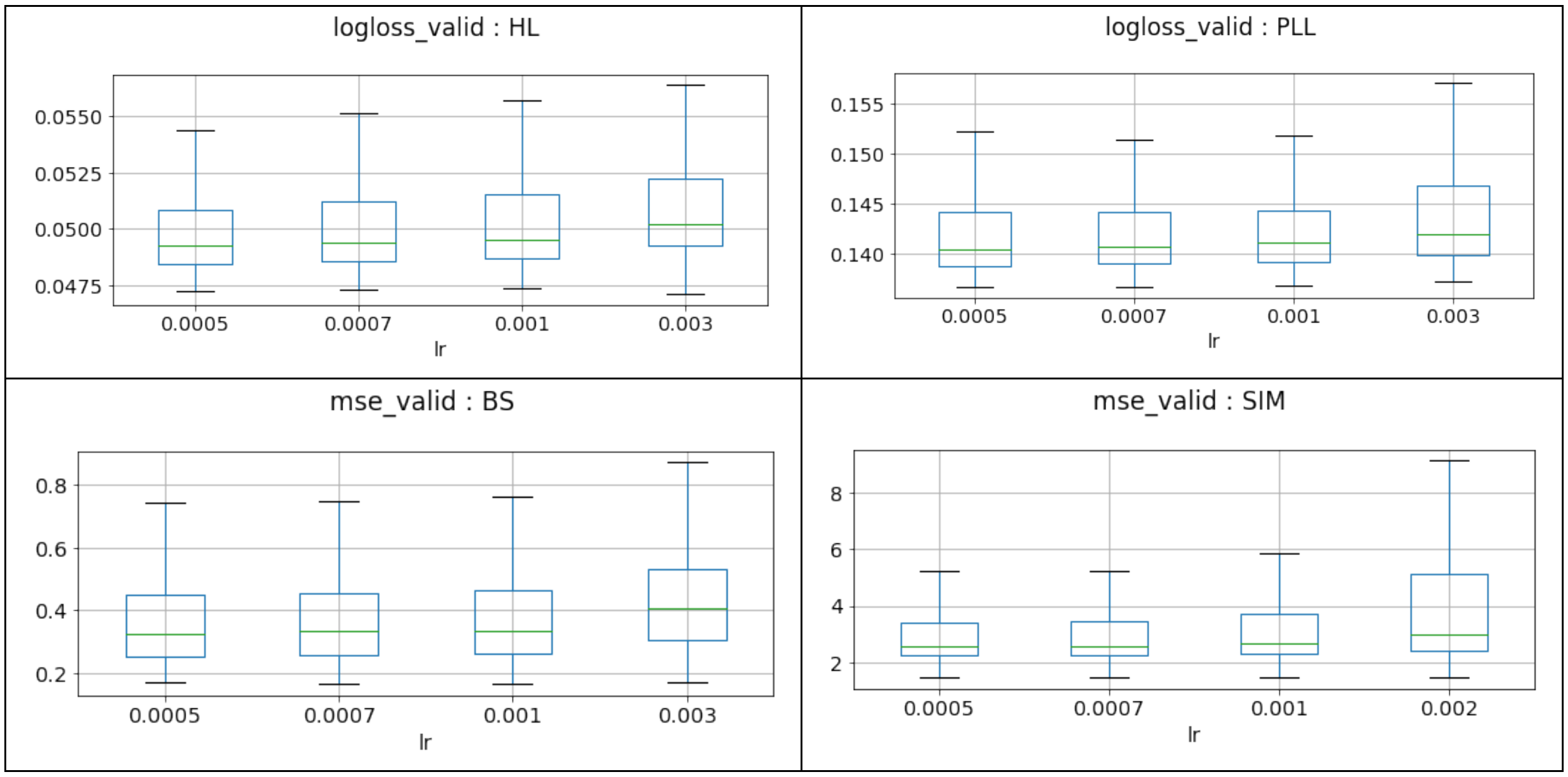

Figure 11 Boxplot of Logloss/MSE for FFNN models grouped by Lr_rate

Contour plots help us visualize the payoff due to interaction between HPs and highlight the optimal region of the models specific to the dataset. Figure 12 shows selected contour plots from our exploratory study across the datasets. In these plots we have picked interaction between the layer sizes, interaction between Layer2 and regularization HP dropout and finally the interaction between two regularization HPs, L1 and dropout. For each pair of HPs (for example, Layer1 vs Layer2) we get the average performance of models for all possible configurations of the selected pair of HPs and obtain the contour plots. For each data set the contour plots are mapped to the same color scale across pairs of HPs for ease of comparison. In case of BS and SIM data, instead of using MSE, we used $R^2$ which is inversely proportional to MSE for constant TSS. As a result, for each of the contour plots lighter region corresponds to high performing models and darker regions correspond to poor performing models. We have observed the following from Figure 12:

i. Optimal regions can differ widely based on the datasets.
ii. Unlike the XGBoost models sometimes pairwise contour plots for FFNN models show disconnected regions of optimal performance (example Layer1 vs Layer2 plot for HL data). However, in our study we did not find any other contour plot that with this particular phenomenon.
iii. Optimal region of regularization HP also depends on presence and size of Layer2 and size of layer 1 as is seen in the plots from second column. For example, in HL data contour plot of Dropout vs Layer2 we see that, in absence of second layer models perform well for the entire range of Dropout as most of the models have sufficiently large Layer1. However, when a second layer is present and the size of this layer is small, a large Dropout can randomly suppress important signals in the data leading to poor performance as shown by the sudden drop in the middle.
iv. The BS and SIM data clearly prefer more complex networks than HL and PLL data. In the contour plot of Dropout vs Layer2 for BS data we see that, in absence of Layer2 models do not prefer any amount of Dropout regularization and clearly prefer a large Layer1(as seen in the Layer1 vs Layer2 plot).

v. We noticed in the ANOVA tables, the strong interactions between the regularization HPs. This is further exemplified in the contour plot of Dropout vs L1 in column three of Figure 12, where it is observed that optimal models tend to prefer only one of these two regularizations. In general model performance can deteriorate drastically as L1 penalty grows.

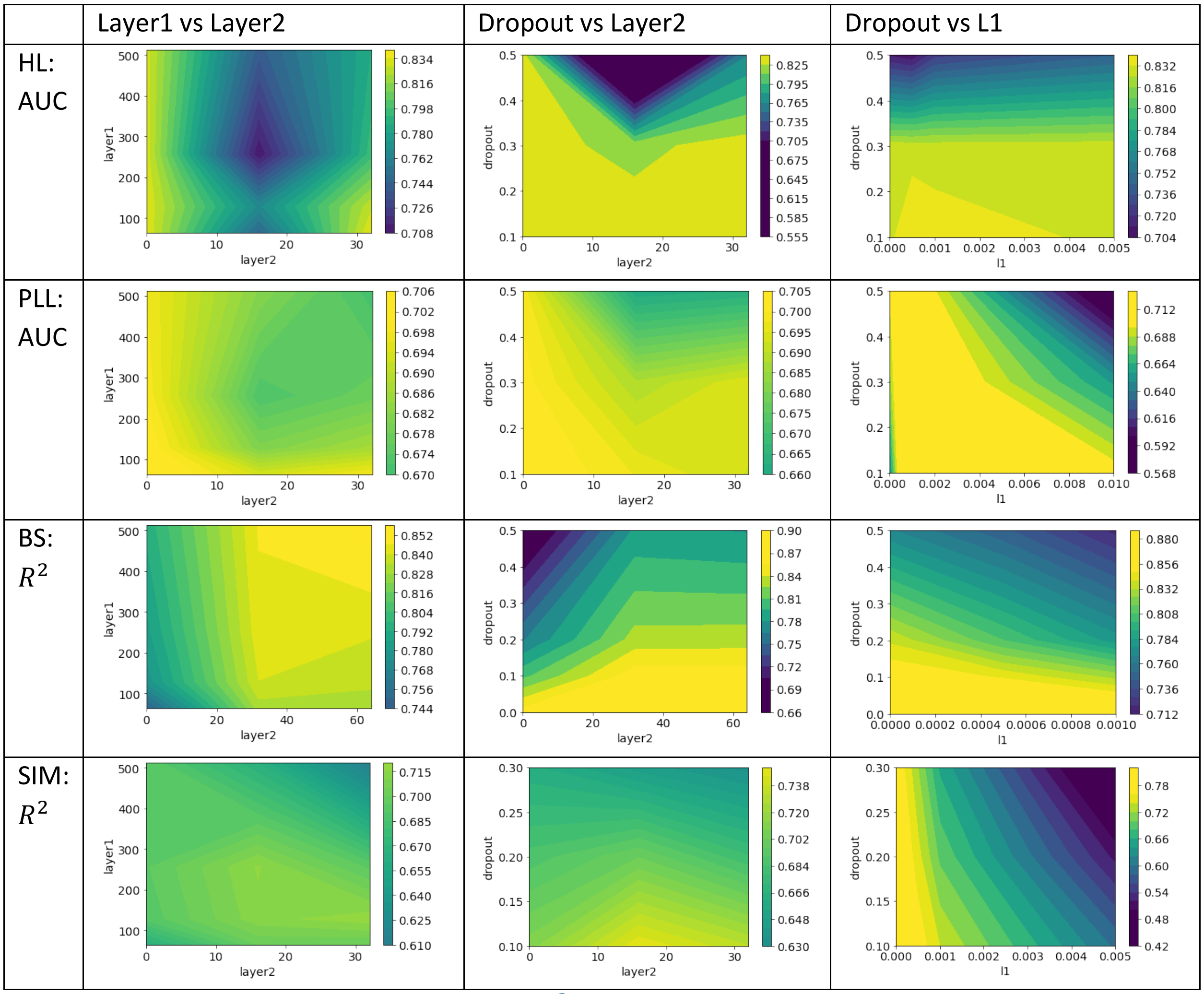

Figure 12: Contour plots of AUC/$\mathbf{R^2}$ for selected pairs of HP for FFNN models

### 6.2. Local Behavior in High-Performing Regions of the Hyper-Parameter Space

Similar to the analysis in Section 4.2 and 5.2, we want to determine existence of additional interactions between HPs in the top models in our search that did not rank high in the ANOVA tables. Again, we looked at the rank correlation between the HPs in the 50 best performing models (see Table 33). In general we see less consistency in correlation across the tables and stronger correlation than that observed in XGBoost models. The correlations also can be very different based on the metric used for model selection which shows each metric may choose different sets of top-ranking models.

Based on these rank correlation matrices, we observe the following:

i. Lack of consistency across datasets – which might be caused by higher order interactions at play.
ii. Layer1 and Layer2 has positive correlation with at least one of the regularizing HP
iii. Regularization HPs are usually negatively correlated suggesting the models under-fit if all regularization HPs are high. Hence overfitting should ideally be controlled by at most one or two regularization HPs
iv. Lr_rate has positive correlation with at least one of the regularizing HP suggesting that when Lr_rate is high, it tends to over-fit and hence needs regularization to control over-fitting.
v. We also observe that in continuous response data models using large Batch_size also tend to use high L2 penalty however this interaction is not observed in binary response data and should be further explored. While further investigation on the underlying reason is beyond the scope of this study, this indicates that even though Batch_size does not have strong direct effect it can still impact local performance through interactions with other HPs.
vi. In the SIM data all 50 models selected had L1 penalty set at 0 and were regularized through L2 and dropout. This suggests that models can be effectively regularized through these two HPs. In the contour plots, it was also evident that slight changes in L1 penalty when combined with any other forms of regularization can lead to drastic change in performance and hence it is easier to regularize models through the use of L2 and dropout.

Table 33 Rank correlation of top 50 FFNN models ranked by Logloss/MSE

<table>
<tr><td>Data</td><td colspan="2"></td></tr>
<tr><td>HL</td><td>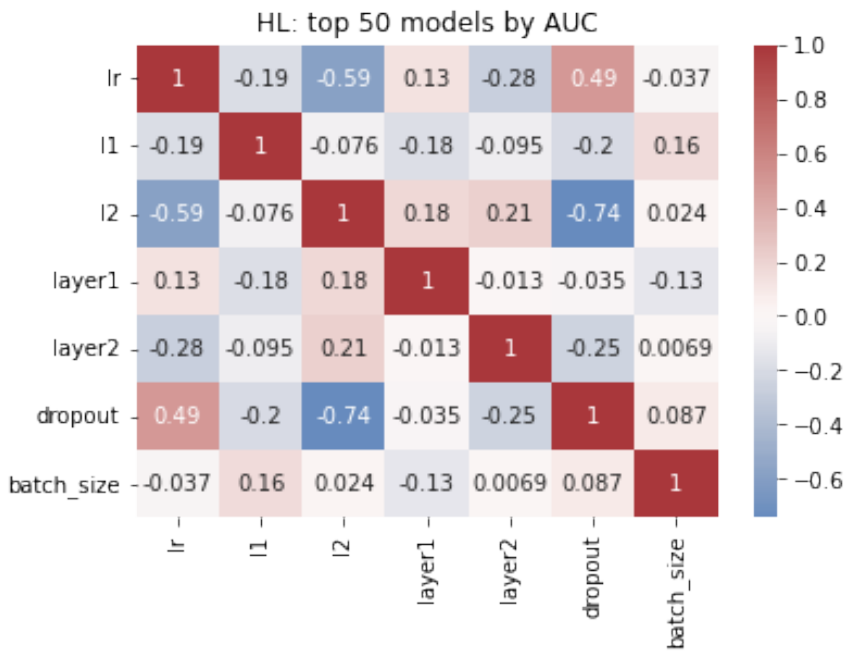
</td><td>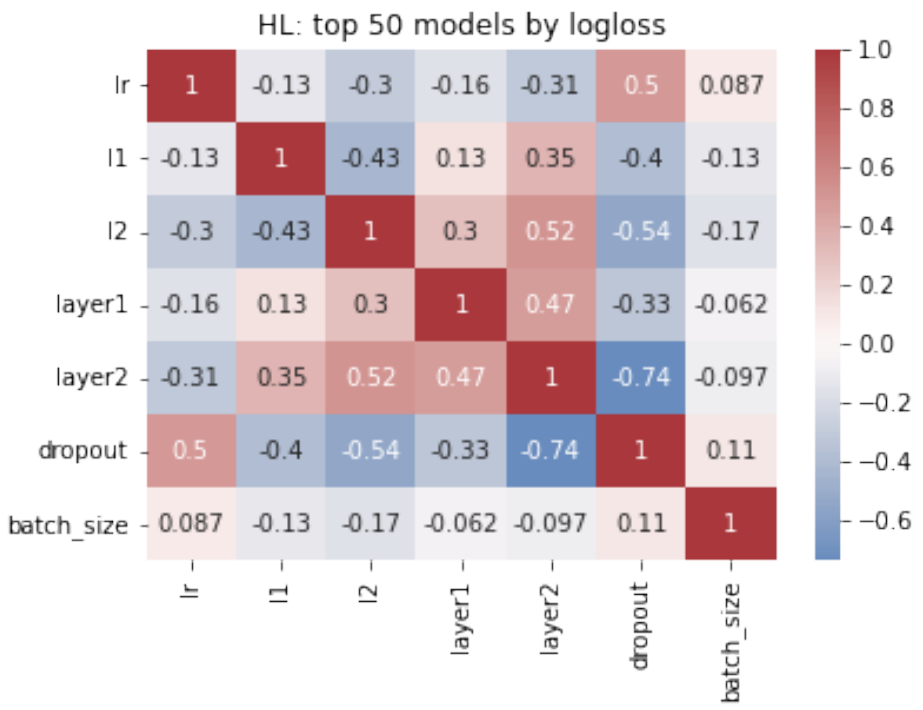
</td></tr>
<tr><td>PLL</td><td>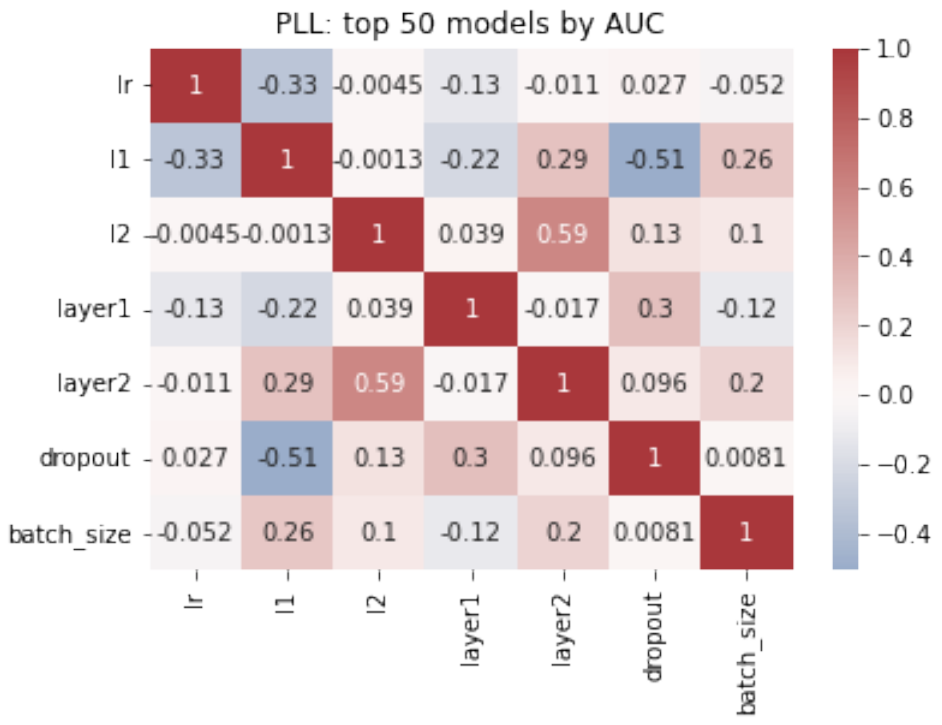
</td><td>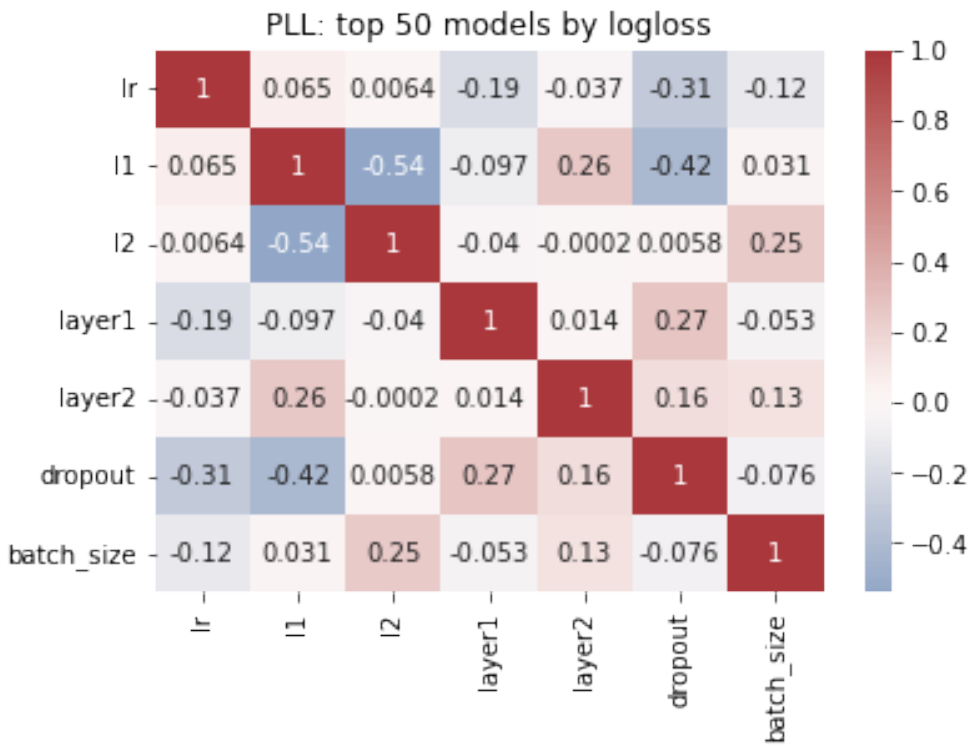
</td></tr>
</table>

<table>
<tr><td>BS</td><td>
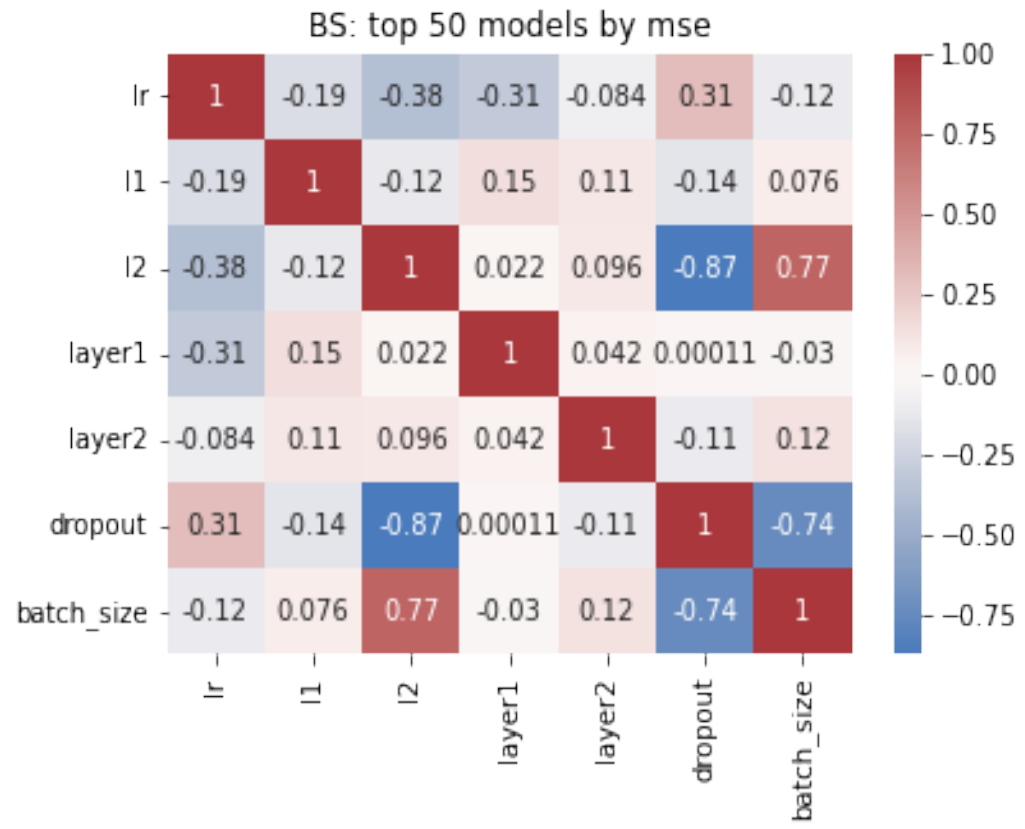

</td></tr>
<tr><td>SIM</td><td>
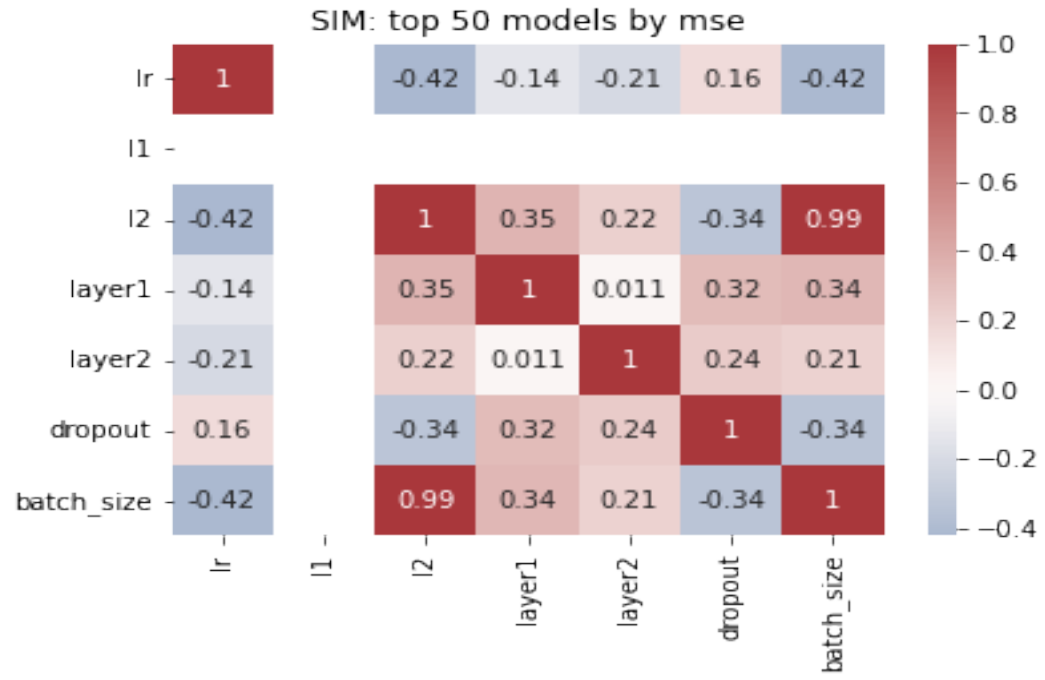

</td></tr>
</table>

### 6.3. Robustness and Overfitting

Figure 13 examines over-fitting in FFNN models when Layer sizes vary while Figure 14 studies over-fitting for the regularization HPs: L1 and L2 penalties and Dropout. The y-axes show the gap statistics and the x-axes show validation Logloss or MSE. The dots are colored according to levels of a particular HP. We see from Figure 13 that, for the most part, very large values of Layer1 (512) and Layer2 (64 or 32) lead to over-fitting although the pattern varies with datasets. We see that there is over-fitting when one or more of the hidden layers are large but not compensated by high regularization. The plots for HL and PLL data in Figure 14 show:

i. Regularization through Dropout alone may not be sufficient and added regularization through L2 or L1 is required to control over-fitting.

<table>
<tr><td></td><td>Layer1</td><td>Layer2</td></tr>
<tr><td>HL</td><td>
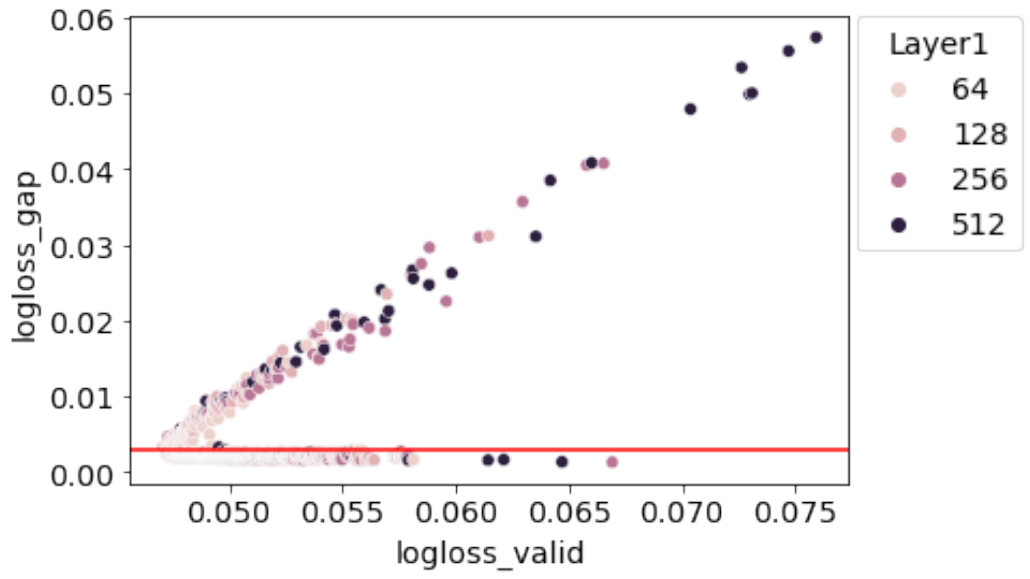

</td><td>
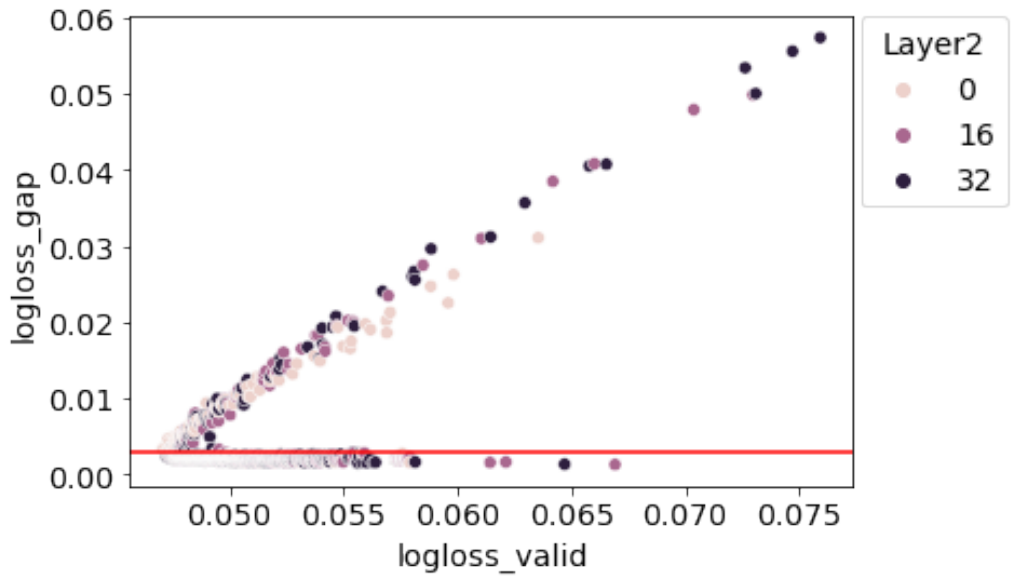

</td></tr>
</table>

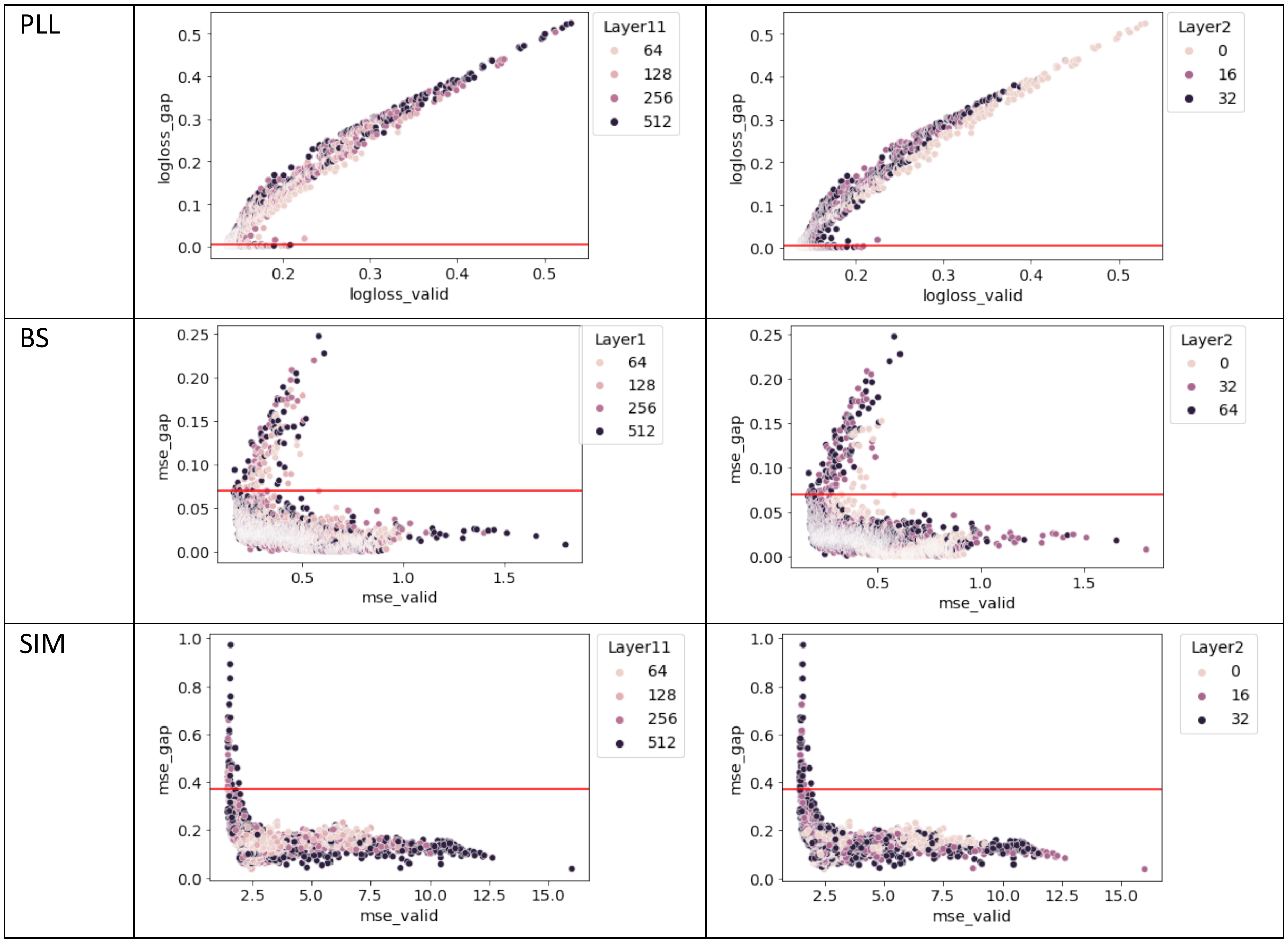

Figure 13: Overfitting in FFNN models:
Exploring with respect to size of network (red line marks gap statistic of optimal model)

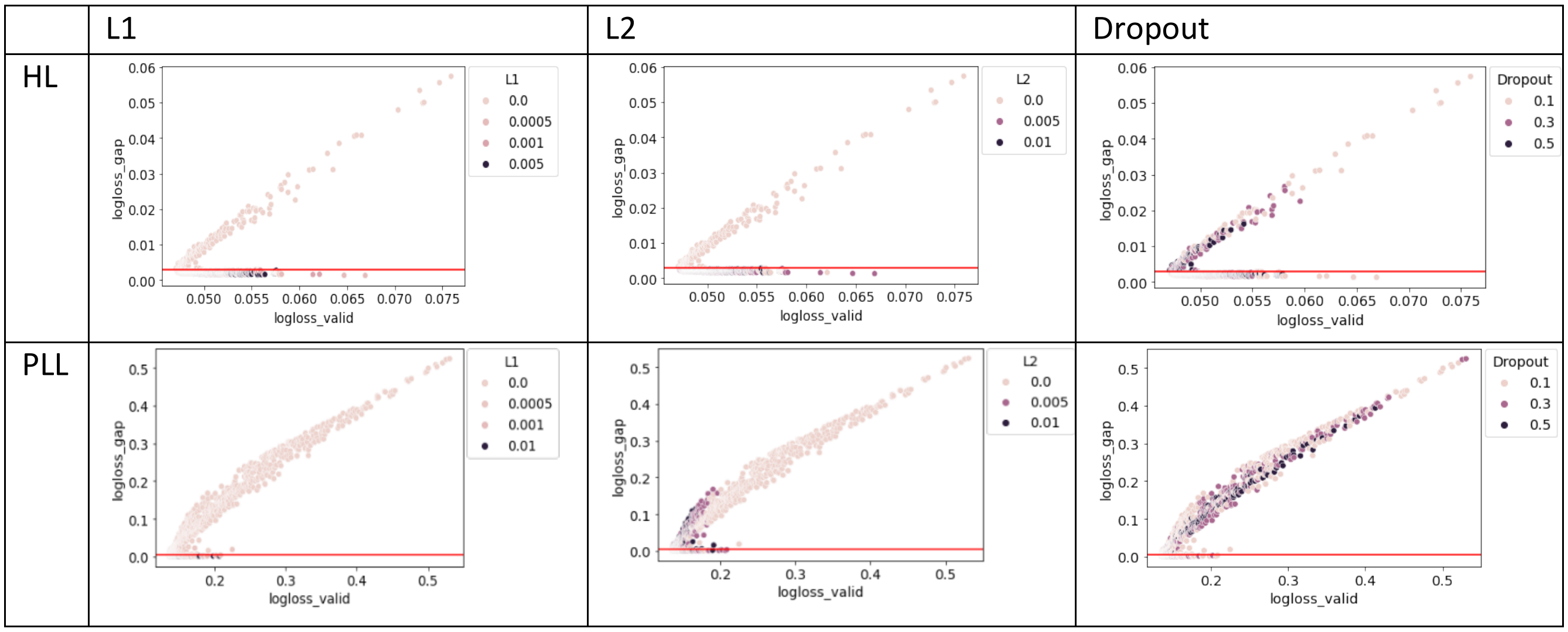

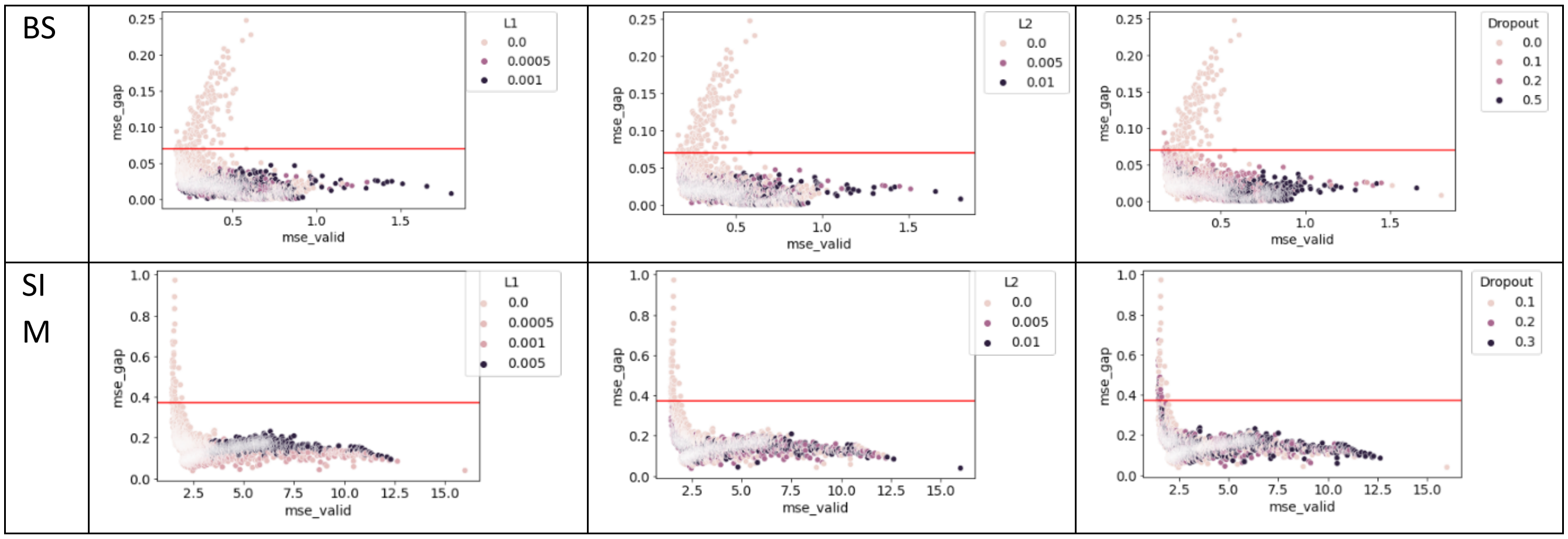

Figure 14: Overfitting in FFNN models:
Exploring with respect to regularization HP (redline marks gap statistic of optimal model)

Table 34: Competing FFNN models for robustness analysis in BS data

| Models | Lr_rate | Layer1 | Layer2 | L1 | L2 | Dropout | Batch_size | Final epoch | MSE_train | Mse_valid | Mse_gap |
|---|---|---|---|---|---|---|---|---|---|---|---|
| Opt | 0.0007 | 512 | 64 | 0 | 0 | 0.2 | 500 | 908 | 0.0968 | 0.1643 | 0.0675 |
| Alt | 0.0007 | 256 | 64 | 0 | 0.005 | 0 | 1000 | 691 | 0.1397 | 0.1749 | 0.0352 |

ii. As with RF and XGB models, we observed that, in general, gap statistics of the optimal models in continuous response data (41.46% for BS and 25.48% for SIM) were higher than those for binary response data (6.38% for HL and 2.92% for PLL). Again, this suggests that the optimal models for the continuous response datasets may have over-fit and hence may be less robust than a competing model with lower training and validation gap.

We further investigated this by comparing it with an alternative, well-performing, model and performing the same type of local perturbation analysis as in Section 4.3 and 5.3 for BS dataset (see Table 34). In this case, however, the optimal model is more robust to perturbations than the competing model despite the higher gap between the training and validation metric. This conclusion is different from the ones from the analysis of RF and XGB models. Thus, one cannot rely exclusively on gap statistics to determine robustness of models.

## 7. AUC vs Logloss

The optimal HP search is based on some performance metric. Usually for continuous regression problems in ML models the popular choice of selection criterion is MSE. The choice is not so clear in case of binary regression problems and modelers can use different metrics to choose their optimal model depending on the business needs. The two popular contenders in this case are AUC and Logloss.

Figure 15 consists of the AUC vs Logloss plots for each of the two binary response datasets for the three algorithms. The good models are concentrated in the top left corner of these plots.

We observe from the plots that the top models based on these two metrics are not always the same.

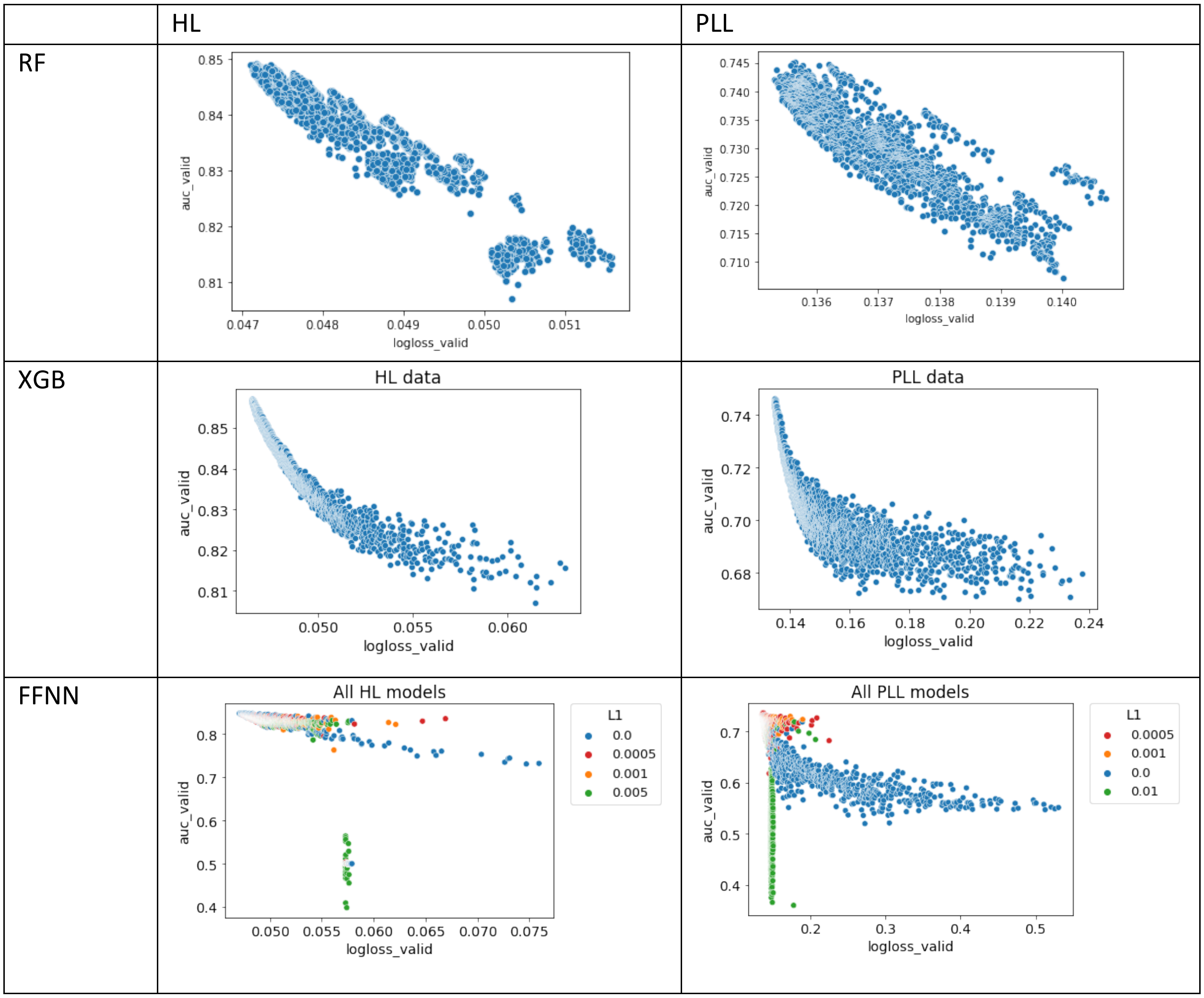

Figure 15: AUC vs Logloss

In RF models we see a strong agreement between AUC and Logloss but they are not exactly collinear leading to a different choice of optimal models. Hence it is possible to find models which perform well with respect to both metrics but the best model chosen is usually different based on the chosen metric.

In XGB models, agreement in the two metrics is strong for high performing models. However exact ranking is different and based on the choice of metric different optimal model may be selected. For the two datasets we looked at, by coincidence both metrics choose same optimal model.

In FFNN models, we observe that though a large percentage of models are in agreement we have a band of models with low AUC values which is not directly reflected in the Logloss values. Due to high regularization these models are equivalent to a near intercept model with no discriminating power and hence the low AUC values. The data imbalance (proportion of label 1 in the validation subset is 1% and 3% respectively for HL and PLL data) forces all estimated probabilities to be low for these near intercept models ($\hat{p}_i \sim 0$). Hence, the Logloss metric measured as $\frac{1}{n}\sum_{i=1}^{n} -(y_i log(p_i) + (1 - y_i)\log(1 - p_i))$ is heavily dominated by the majority label in the data (in our case 0) which prevents the Logloss value from growing beyond a certain point for these near intercept models.

In Figure 16, we look at AUC vs Logloss for the top 500 models selected by ranking the models with respect to Logloss. There is a general agreement between the metrics among the top models, however the two metrics may not select the same model as optimal. We observe that the top models in both the scenarios exclude models with the top L1 regularization value, thus again confirming the observation that FFNN models react poorly to high L1 penalty if the network is not exceptionally wide or deep and already regularized with other HPs.

<table>
<tr><td></td><td>Top 500 models (ranked by Logloss)</td></tr>
<tr><td>HL</td><td>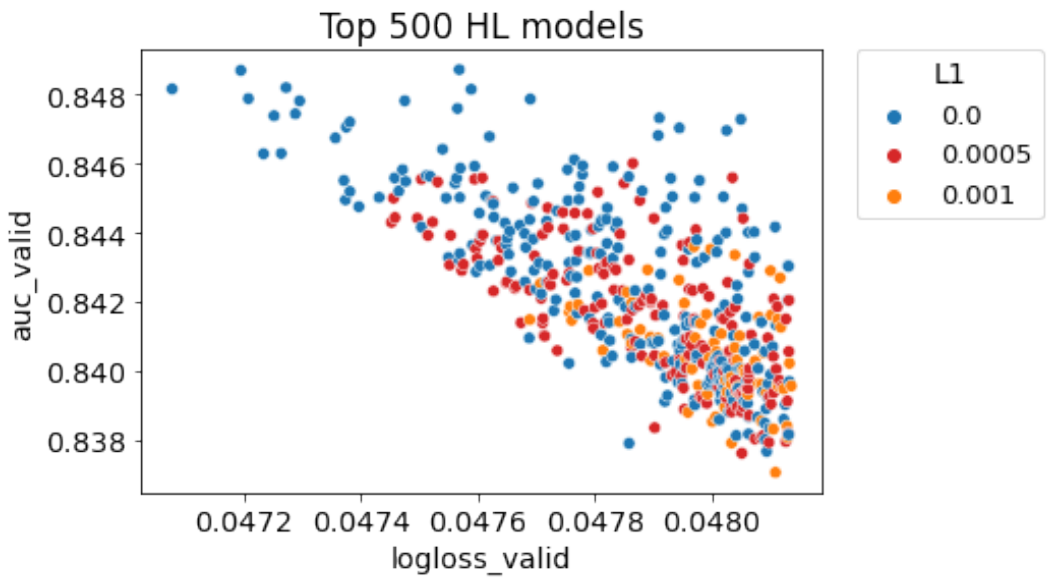
</td></tr>
<tr><td>PLL</td><td>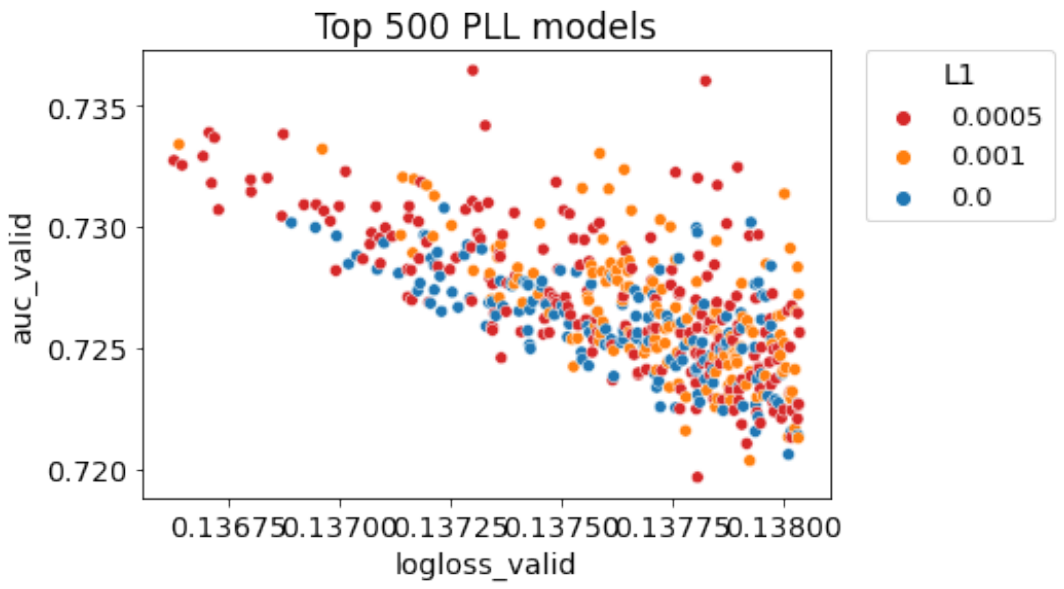
</td></tr>
</table>

Figure 16: AUC vs Logloss in top 500 FFNN models

## 8. Concluding Remarks

This empirical investigation provides useful insights on how the performance of fitted models vary with configurations of HPs, including global and location behavior. Not surprisingly, the interactions can lead to multiple models with similar performances. As we have demonstrated, this understanding can facilitate in developing multi-stage search strategies to reduce the search space and computational burden. The views expressed in the paper are those of the authors and do not represent the view of Wells Fargo.

## A. Appendix

### a. Results of Two-Stage Strategy for RF Models

Table 35: Comparison of reduced strategy with extensive grid search in RF models
Red line marks performances of r-opt model

| Data | metric | Boxplot of performances of all models | Percentage of models having higher performance | Relative decrease in performance |
|---|---|---|---|---|

<table>
<tr><td rowspan="2">HL</td><td>AUC</td><td>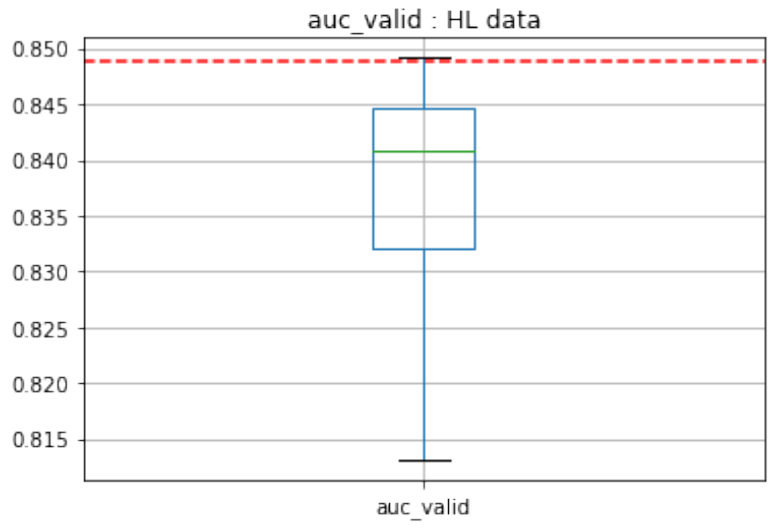
</td><td>0.17%</td><td>0.05%</td></tr>
<tr><td>Logloss</td><td>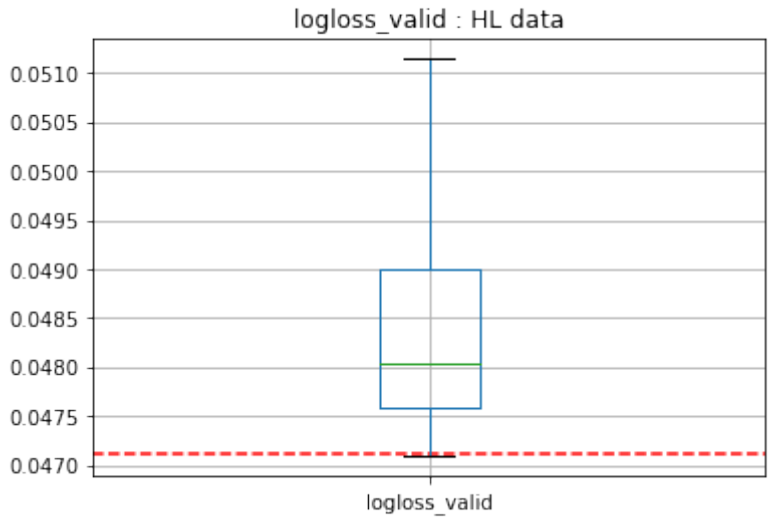
</td><td>0.03%</td><td>0.06%</td></tr>
<tr><td rowspan="2">PLL</td><td>AUC</td><td>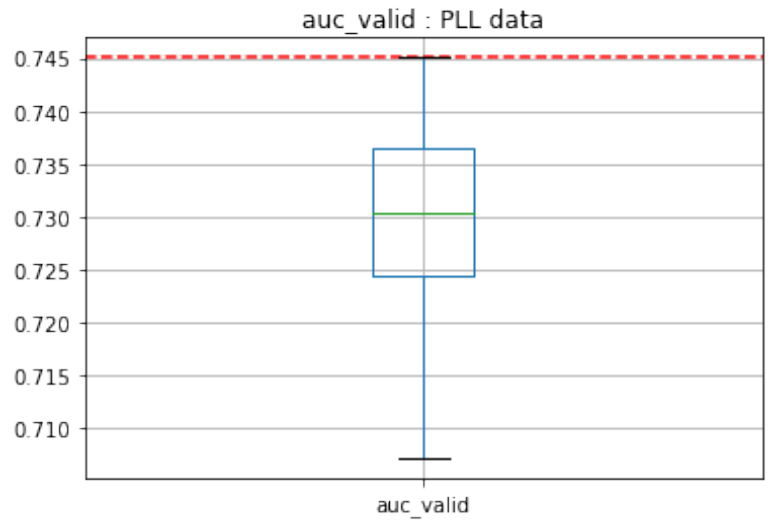
</td><td>0.2%</td><td>0.07%</td></tr>
<tr><td>Logloss</td><td>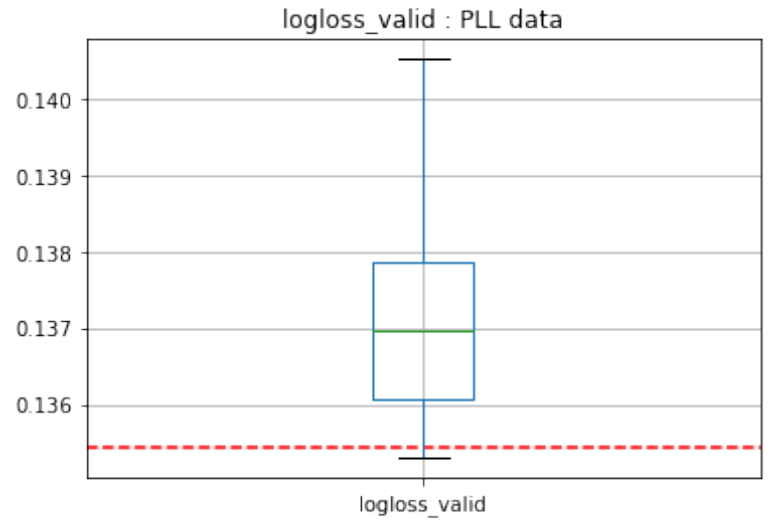
</td><td>0.92%</td><td>0.1%</td></tr>
<tr><td>BS</td><td>MSE</td><td>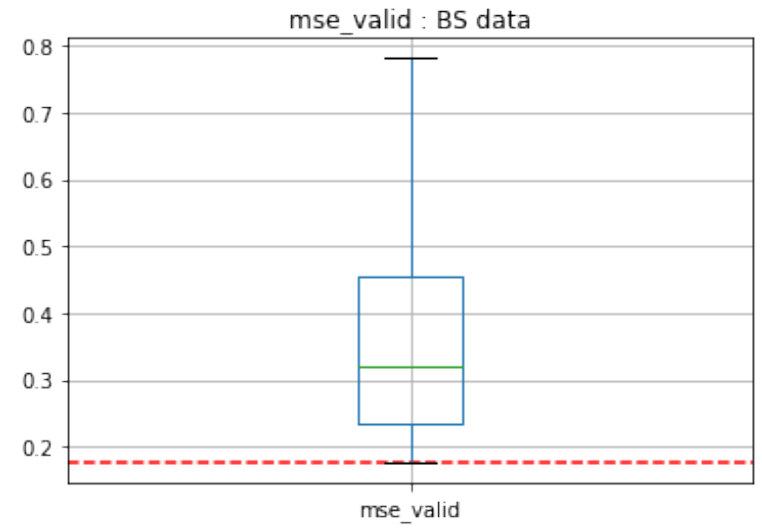
</td><td>best</td><td>0%</td></tr>
</table>

### b. Results of Two-Stage Strategy for XGB Models

Table 36: Comparison of reduced strategy with extensive grid search in XGB models (Red line marks the performances of r-opt model)

| Data | Metric | Boxplot of performances of all 6,250 models for each dataset | Percentage of models having higher performance) | Relative change in performance 100 |
|---|---|---|---|---|
| HL | AUC | 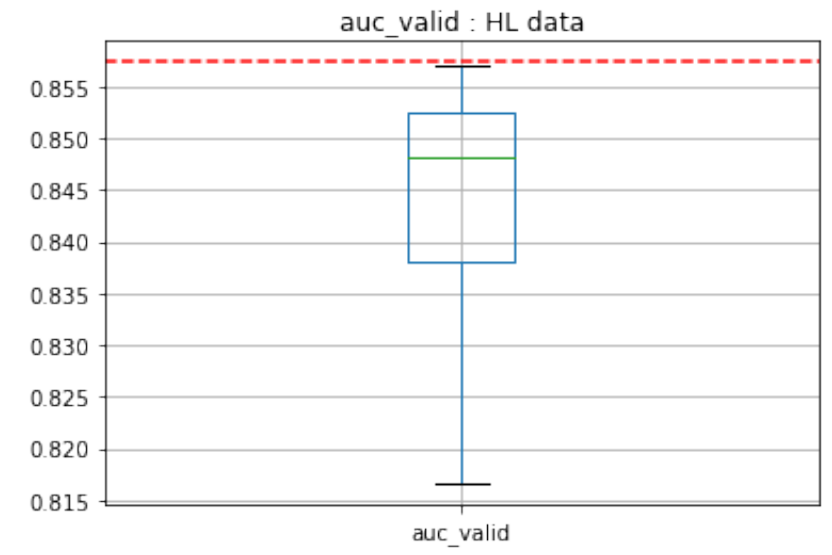 | best | 0.05% |
| | Logloss | 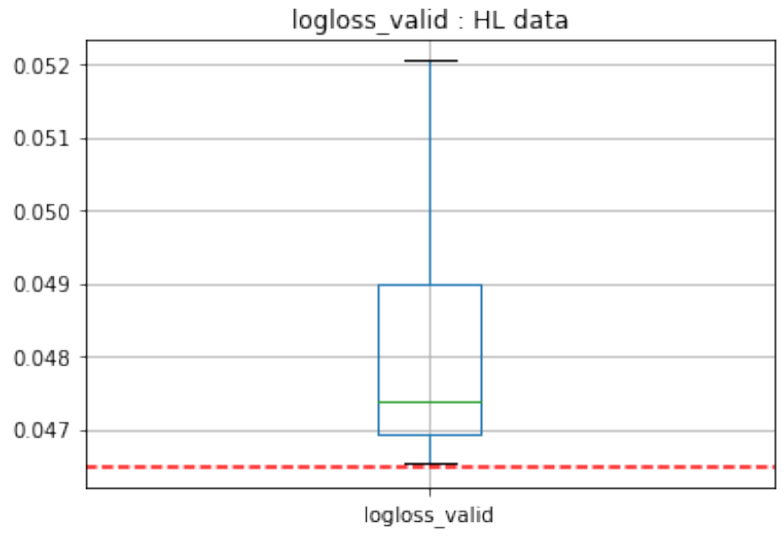 | best | 0.02% |
| PLL | AUC | 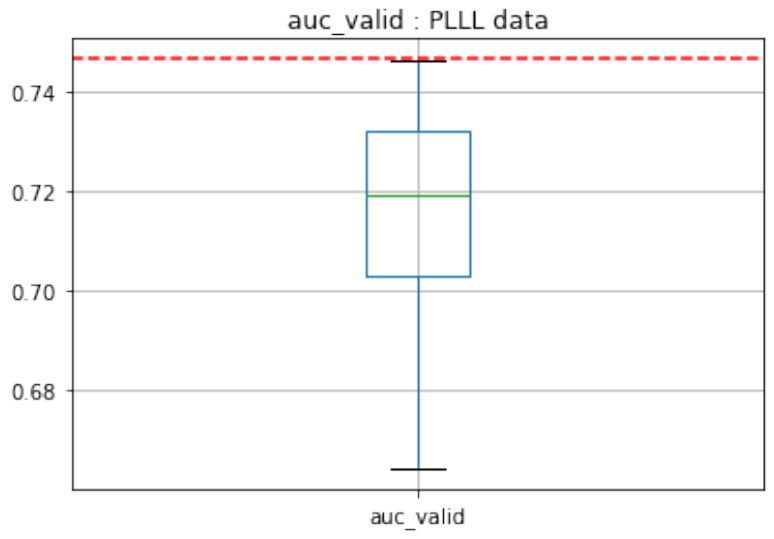 | best | 0.01% |
| | Logloss | 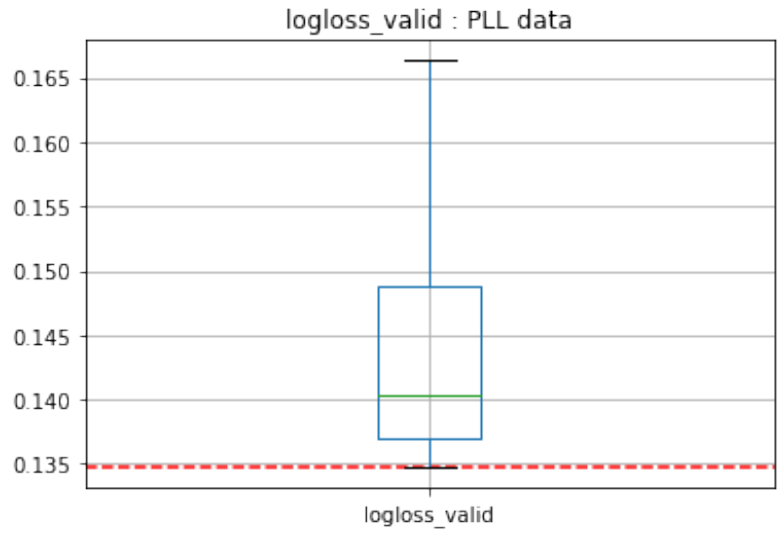 | 0.016% | 0.03% |
| BS | MSE | 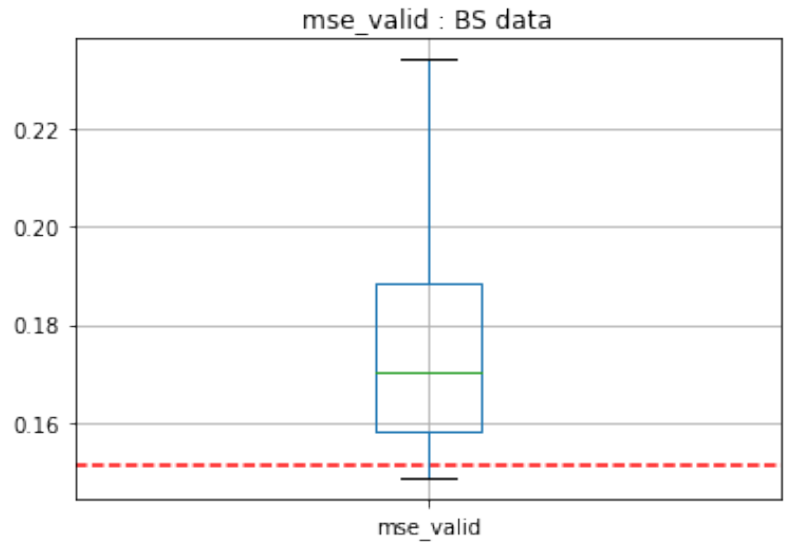 | 1.47% | 1.9% |

### c. Results of Two-Stage Strategy for FFNN Models

Table 37: Comparison of reduced strategy with extensive grid search in FFNN models (Red line marks the performances of r-opt model)

| Data | Metric | Boxplot of performances of all models | Percentage of models having higher performance | Relative decrease in performance by adopting two stage strategy |
|---|---|---|---|---|
| HL | AUC | 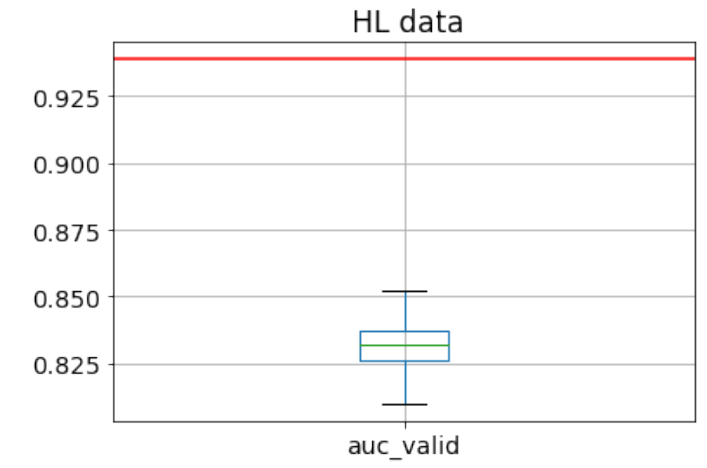 | best | 0% |
| | Logloss | 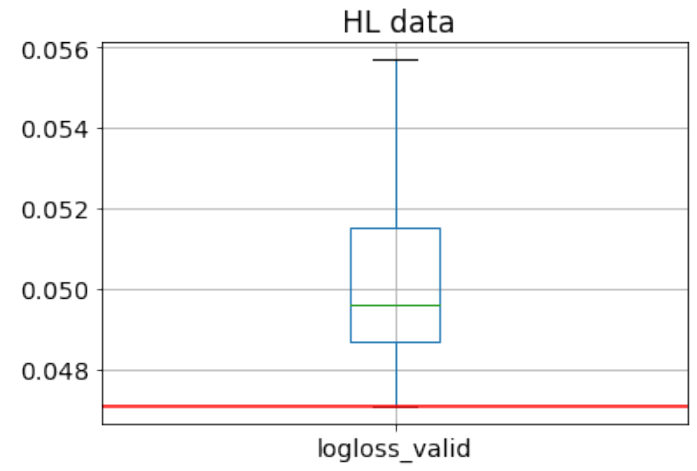 | best | 0% |
| PLL | AUC | 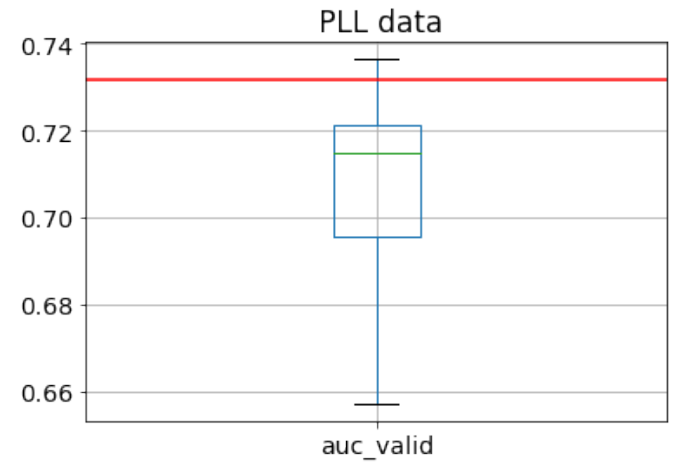 | 0.42% | 0.66% |
| | Logloss | 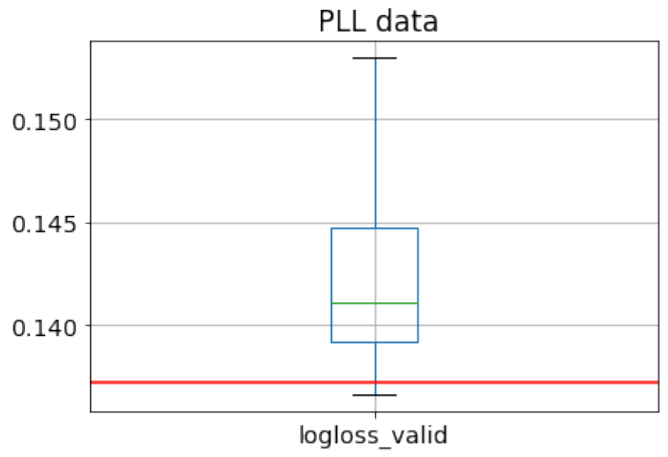 | 0.95% | 0.44% |
| BS | MSE | 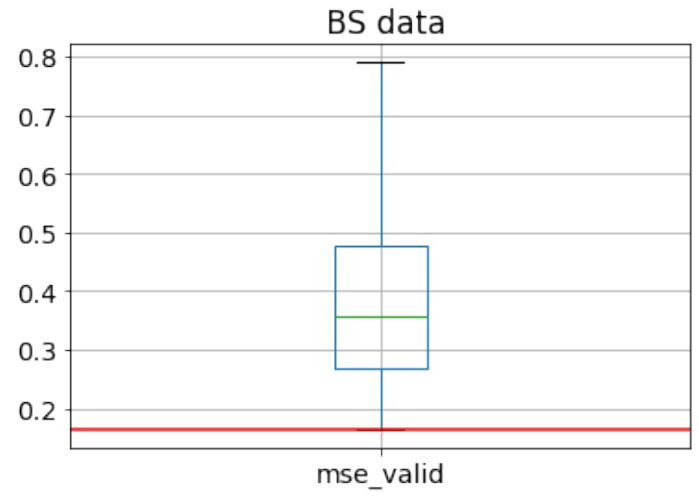 | best | 0% |
| SIM | MSE | 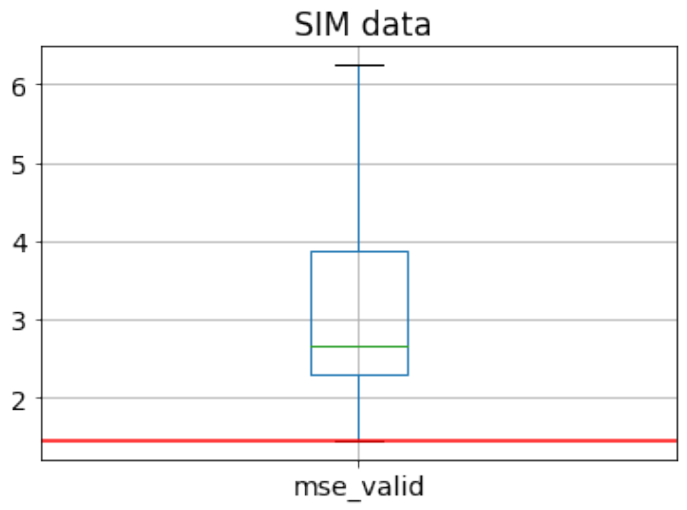  | best | 0% |

### d. Extension Study on Minimum Child Weight and Column Sample Rate

As per the XGBoost documentation (XGBoost Parameters, 2022), **minimum child weight** (Min_child_weight) is the minimum sum of instance weight needed in a child. If a tree partition results in a leaf node with the sum of instance weight less than this parameter then the building process gives up further partitioning. In a linear regression task this amounts to the minimum number of instances needed in each node. Tuning this Hyper-Parameter prevents Trees to grow too deep and can help control over-fitting.

On the other hand the Hyper-Parameter **column sample by tree** (colsample_bytree) is the subsample ratio of columns when constructing each tree and helps to control correlation between the trees, which in turn can prevent over-fitting.

We performed a small side study on the effectiveness of these two Hyper-Parameters and if they can provide additional assistance to the five HPs already being tuned. In order to explore this we chose two models, one which is optimal in our grid search and a model that has seriously over-fit due to lack of regularization.

We performed additional tuning on these models with respect to Min_child_weight and colsample_bytree and report the relative gain in Logloss/MSE for the respective models. The grid for min child weight looks at geometrical progression of values from 1 to $\sqrt{n}$ where $n$ is the number of observations in the training frame (weight of each instance is 1 in our data) and column sample grid is chosen as geometrically spaced values from 0.05 to 1 where these values denote the proportion of columns sampled per tree.

The grid for the two HPs are listed in Table 38

Table 38: Grid for tuning Min_child_weight and Colsample_bytree by tree

| | Min_child_weight | Colsample_bytree |
|---|---|---|
| HL (n ~ 333K) | 1., 2., 4., 8., 16., 34., 69., 140., 284 ., 577. | 0.05, 0.07, 0.1, 0.14, 0.19, 0.26, 0.37, 0.51, 0.72, 1 |
| PLL (n ~ 30K) | 1., 3., 5., 10., 18., 32., 58., 105., 18 8. | 0.05, 0.07, 0.1, 0.14, 0.19, 0.26, 0.37, 0.51, 0.72, 1 |
| BS (n ~ 10K) | 1., 2., 4., 7., 12., 21., 35., 58., 96. | 0.05, 0.07, 0.1, 0.14, 0.19, 0.26, 0.37, 0.51, 0.72, 1 |

Column 3 in list the optimal model and an over-fit model from our extensive grid before tuning of Min_child_weight and colsample_bytree. The fourth column shows the modified model when Min_child_weight and colsample_bytree are tuned holding the other five HPs fixed at given value. The fifth column shows the resultant relative change in the metric in the validation frame along with the relative change in the corresponding gap statistic. The results show that we do not gain much by additional tuning on the optimal model, however when the over-fit model is tuned there is indeed a large change in the metric and the gap statistic also lowers considerably, demonstrating the regularization ability of these two additional HPs.

Table 39: Results from tuning Min_child_weight and Colsample_bytree by tree in XGB models

Orange – additional HPs tuned in extended search, Grey – Metrics for evaluation

| Data | | Models before tuning Min_child_weight and colsample_bytree | | Models post tuning | | Relative change in metric $\frac{\lvert metric_{pre} - metric_{post}\rvert}{metric_{post}} \times 100$ | |
|---|---|---|---|---|---|---|---|
| HL | Optimal (M1) | Depth | 4 | Depth | 4 | Logloss_valid: | 0.1% |
| | | Trees: | 500 | Trees: | 500 | Logloss_gap: | 8.3% |
| | | Lr_rate | 0.052 | Lr_rate | 0.052 | | |
| | | L1: | 10 | L1: | 10 | | |
| | | L2: | 2.15 | L2: | 2.15 | | |
| | | Min_child_weight | 1 | Min_child_weight | 1 | | |
| | | Colsample_bytree | 1 | Colsample_bytree | 0.51 | | |
| | | Logloss_valid: | 0.04658 | Logloss_valid: | 0.04652 | | |
| | | Logloss_gap: | 0.0065 | Logloss_gap: | 0.0060 | | |
| | Over-fit (M2) | Depth | 4 | Depth | 4 | Logloss_valid: | 4% |
| | | Trees: | 500 | Trees: | 500 | Logloss_gap: | 267% |
| | | Lr_rate | 0.2 | Lr_rate | 0.2 | | |
| | | L1: | 0.464 | L1: | 0.464 | | |
| | | L2: | 0.1 | L2: | 0.1 | | |
| | | Min_child_weight | 1 | Min_child_weight | 284 | | |
| | | Colsample_bytree | 1 | Colsample_bytree | 0.26 | | |
| | | Logloss_valid: | 0.049 | Logloss_valid: | 0.047 | | |
| | | Logloss_gap: | 0.024 | Logloss_gap: | 0.007 | | |
| PLL | Optimal (M1) | Depth | 3 | Depth | 3 | Logloss_valid: | 0.2% |
| | | Trees: | 300 | Trees: | 300 | Logloss_gap: | 37.5% |
| | | Lr_rate | 0.052 | Lr_rate | 0.052 | | |
| | | L1: | 10 | L1: | 10 | | |
| | | L2: | 10 | L2: | 10 | | |
| | | Min_child_weight | 1 | Min_child_weight | 10 | | |
| | | Colsample_bytree | 1 | Colsample_bytree | 0.14 | | |
| | | Logloss_valid: | 0.1348 | Logloss_valid: | 0.1345 | | |
| | | Logloss_gap: | 0.011 | Logloss_gap: | 0.008 | | |
| | Over-fit (M2) | Depth | 3 | Depth | 3 | Logloss_valid: | 8.9% |
| | | Trees: | 500 | Trees: | 500 | Logloss_gap: | 90% |
| | | Lr_rate | 0.2 | Lr_rate | 0.2 | | |
| | | L1: | 0.1 | L1: | 0.1 | | |
| | | L2: | 0.1 | L2: | 0.1 | | |
| | | Min_child_weight | 1 | Min_child_weight | 188 | | |
| | | Colsample_bytree | 1 | Colsample_bytree | 0.05 | | |
| | | Logloss_valid: | 0.149 | Logloss_valid: | 0.137 | | |
| | | Logloss_gap: | 0.09 | Logloss_gap: | 0.02 | | |

<table>
<tr>
<td>BS</td>
<td>Optimal (M1)</td>
<td><table>
<tr><td>Depth</td><td>6</td></tr>
<tr><td>Trees:</td><td>400</td></tr>
<tr><td>Lr_rate</td><td>0.178</td></tr>
<tr><td>L1:</td><td>0</td></tr>
<tr><td>L2:</td><td>10</td></tr>
<tr><td>Min_child_weight</td><td>1</td></tr>
<tr><td>Colsample_bytree</td><td>1</td></tr>
<tr><td>MSE_valid:</td><td>0.149</td></tr>
<tr><td>MSE_gap:</td><td>0.1</td></tr>
</table></td>
<td><table>
<tr><td>Depth</td><td>6</td></tr>
<tr><td>Trees:</td><td>400</td></tr>
<tr><td>Lr_rate</td><td>0.178</td></tr>
<tr><td>L1:</td><td>0</td></tr>
<tr><td>L2:</td><td>10</td></tr>
<tr><td>Min_child_weight</td><td>7</td></tr>
<tr><td>Colsample_bytree</td><td>0.7</td></tr>
<tr><td>MSE_valid:</td><td>0.149</td></tr>
<tr><td>MSE_gap:</td><td>0.09</td></tr>
</table></td>
<td><table>
<tr><td>MSE_valid:</td><td>2%</td></tr>
<tr><td>MSE_gap:</td><td>15%</td></tr>
</table></td>
</tr>
<tr>
<td></td>
<td>Over-fit (M2)</td>
<td><table>
<tr><td>Depth</td><td>6</td></tr>
<tr><td>Trees:</td><td>500</td></tr>
<tr><td>Lr_rate</td><td>0.2</td></tr>
<tr><td>L1:</td><td>0</td></tr>
<tr><td>L2:</td><td>0</td></tr>
<tr><td>Min_child_weight</td><td>1</td></tr>
<tr><td>Colsample_bytree</td><td>1</td></tr>
<tr><td>Logloss_valid:</td><td>0.165</td></tr>
<tr><td>Logloss_gap:</td><td>0.148</td></tr>
</table></td>
<td><table>
<tr><td>Depth</td><td>6</td></tr>
<tr><td>Trees:</td><td>500</td></tr>
<tr><td>Lr_rate</td><td>0.2</td></tr>
<tr><td>L1:</td><td>0</td></tr>
<tr><td>L2:</td><td>0</td></tr>
<tr><td>Min_child_weight</td><td>96</td></tr>
<tr><td>Colsample_bytree</td><td>1</td></tr>
<tr><td>Logloss_valid:</td><td>0.154</td></tr>
<tr><td>Logloss_gap:</td><td>0.08</td></tr>
</table></td>
<td><table>
<tr><td>MSE_valid:</td><td>7%</td></tr>
<tr><td>MSE_gap:</td><td>85%</td></tr>
</table></td>
</tr>
</table>